\documentclass{article}

\usepackage[accepted]{arxiv}
\usepackage{float}
\usepackage{graphicx} 
\usepackage{amsmath}
\usepackage{amsfonts}
\usepackage{glossaries}
\usepackage{adjustbox}
\usepackage{xcolor}
\usepackage{xspace}
\usepackage{booktabs}
\usepackage{multicol}
\usepackage{multirow} 
\usepackage{hyperref}

\hypersetup{
    colorlinks,
    linkcolor={red!50!black},
    citecolor={blue!50!black},
    urlcolor={blue!80!black}
}

\newcommand{\x}{\textbf{x} }
\newcommand{\xr}{\tilde{\textbf{x}} }
\newcommand{\xrs}{\tilde{x} }

\newacronym{ffn}{FFN}{Feed-Forward Network}
\newacronym{ai}{AI}{Artificial Intelligence}
\newacronym{dl}{DL}{Deep Learning}
\newacronym{dnn}{DNN}{Deep Neural Network}
\newacronym{ood}{OOD}{Out Of Distribution}
\newacronym{xai}{XAI}{eXplainable Artificial Intelligence}
\newacronym{crp}{CRP}{Concept Relevance Propagation}
\newacronym{cav}{CAV}{Concept Activation Vector}
\newacronym{ml}{ML}{Machine Learning}
\newacronym{auc}{AUC}{Area Under the Curve}
\newacronym{se}{SE}{Standard Error}
\newacronym{llm}{LLM}{Large Language Model}
\newacronym{lrp}{LRP}{Layer-wise Relevance Propagation}
\newacronym{iou}{IoU}{Intersection over Union}
\newacronym{amax}{ActMax}{Activation Maximization}
\newacronym{cnn}{CNN}{Convolutional Neural Network}
\newacronym{ours}{AttnLRP}{LRP for Attention}
\newacronym{vit}{ViT}{Vision Transformer}
\newacronym{moe}{MoE}{Mixture of Experts}

\newlength\myHeight 
\newlength\myWidth

\makeatletter
\DeclareRobustCommand\onedot{\futurelet\@let@token\@onedot}
\def\@onedot{\ifx\@let@token.\else.\null\fi\xspace}

\def\eg{\emph{e.g}\onedot} 
\def\ie{\emph{i.e}\onedot}

\def\wrt{w.r.t\onedot} 
 
\title{AttnLRP: Attention-Aware Layer-Wise Relevance Propagation \\for Transformers} 

\author{Reduan Achtibat$^1$ \and
Sayed Mohammad Vakilzadeh Hatefi$^1$ \and
Maximilian Dreyer$^{1}$ \and
Aakriti Jain$^{1}$ \and
Thomas Wiegand$^{1,2,3}$ \and
Sebastian Lapuschkin$^{1,\dagger}$ \and
Wojciech Samek$^{1,2,3,\dagger}$
}

\date{
\footnotesize
$^1$ Fraunhofer Heinrich-Hertz-Institute, 10587 Berlin, Germany\\
$^2$ Technische Universität Berlin, 10587 Berlin, Germany\\
$^3$ BIFOLD – Berlin Institute for the Foundations of Learning and Data, 10587 Berlin, Germany\\
$^\dagger$ corresponding authors: \texttt{\{wojciech.samek,sebastian.lapuschkin\}@hhi.fraunhofer.de}
} 

\begin{document}

\maketitle

\begin{abstract}
Large Language Models are prone to biased predictions and hallucinations, underlining the paramount importance of understanding their model-internal reasoning process.
However, achieving faithful attributions for the entirety of a black-box transformer model and maintaining computational efficiency is an unsolved challenge.
By extending the Layer-wise Relevance Propagation attribution method to handle attention layers, we address these challenges effectively. While partial solutions exist, our method is the first to faithfully and holistically attribute not only input but also latent representations of transformer models with the computational efficiency similar to a single backward pass. Through extensive evaluations against existing methods on LLaMa 2, Mixtral 8x7b, Flan-T5 and vision transformer architectures, we demonstrate that our proposed approach surpasses alternative methods in terms of faithfulness and enables the understanding of latent representations, opening up the door for concept-based explanations.
We provide an LRP library at \url{https://github.com/rachtibat/LRP-eXplains-Transformers}.
\end{abstract}

\section{Introduction}
The attention mechanism~\cite{vaswani2017attention} became an essential component of large transformers due to its unique ability to handle multimodality and to scale to billions of training samples. While these models demonstrate impressive performance in text and image generation, they are prone to biased predictions and hallucinations~\cite{huang2023survey}, which hamper their widespread adoption.

To overcome these limitations, it is crucial to understand the latent reasoning process of transformer models. Researchers started using the attention mechanism of transformers as a means to understand how input tokens interact with each other. Attention maps contain rich information about the data distribution~\cite{clark2019does, caron2021emerging}, even allowing for image data segmentation. However, attention, by itself, is inadequate for comprehending the full spectrum of model behavior~\cite{wiegreffe2019attention}. Similar to latent activations, attention is not class-specific and solely provides an explanation for the softmax output (in attention layers) while disregarding other model components. Recent works~\cite{geva2021transformer, dai2022knowledge} have in fact discovered that factual knowledge in \glspl{llm} is stored in \gls{ffn} neurons, separate from attention layers. 
\begin{figure}[t]
    \centering
    \includegraphics[width=0.99\linewidth]{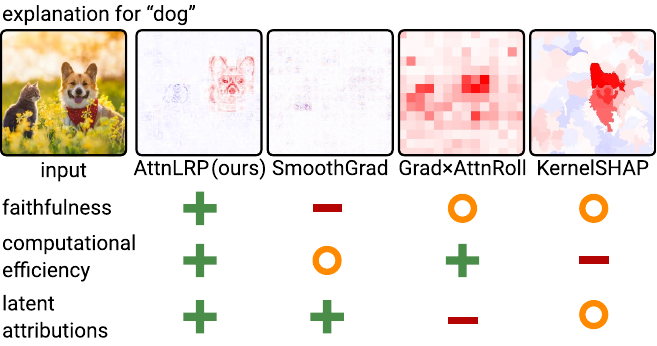}
    \caption{By optimizing LRP for transformer-based architectures, our LRP variant outperforms other state-of-the-art methods in terms of explanation faithfulness and computational efficiency. We further are able to explain latent neurons inside and outside the attention module, allowing us to interact with the model. A more detailed discussion on the differences between \gls{ours} and other LRP variants can be found in Appendix \ref{app:difference}. Heatmaps for other methods are illustrated in Appendix Figure \ref{fig:vit_heatmaps}. Legend: highly ($+$), semi- ($\circ$), not ($-$) suited. Credit: Nataba/iStock.}
    \label{fig:overview}
\end{figure}
Further, attention-based attribution methods such as rollout~\cite{abnar2020quantifying, chefer2021generic} result in checkerboard artifacts, as visible in  Figure~\ref{fig:overview} for a \gls{vit}.
Researchers thus have turned to model-agnostic approaches that aim to provide a holistic explanation of the model's behavior~\cite{miglani2023using},
including, \eg, perturbation and gradient-based methods.

Methods based on feature perturbation require excessive amounts of compute time (and energy), and in order to access latent attributions they require performing perturbations at each layer separately, resulting in further exponential cost increase. This makes their application economically infeasible, especially for large architectures. In contrast, gradient-based methods benefit from the chain-rule in automatic differentiation and can produce latent attributions for all layers in a single backward pass.
While prominent gradient-based methods, \eg Input $\times$ Gradient~\cite{simonyan2014deep}, are highly efficient, they suffer from noisy gradients and low faithfulness, as evaluated in Section~\ref{experiments:evaluation}.

Another option is to take advantage of the versatility of \emph{rule-based} backpropagation methods, such as \gls{lrp}. These methods allow for the customization of propagation rules to accommodate novel operations, allowing for more faithful explanations and requiring only a single backward pass.
As thoroughly discussed in Appendix \ref{app:difference}, all previous attempts to apply LRP to transformers reused standard LRP rules \cite{ding2017visualizing, voita2020analyzing, chefer2021transformer, ali2022xai}. However, transformer architectures include several functions for which standard LRP rules do not adequately apply, such as softmax, bi-linear (matrix) multiplication (\eg query-key multiplication) and layer normalization. Additionally, the routing networks in \gls{moe} models \cite{fedus2022review} present notable challenges due to their combination of these functions. As a result, other previous attempts result in either numerical instabilities or low faithfulness.

Our method represents a significant breakthrough in handling the attribution problem in transformer architectures by enabling an accurate attribution flow through non-linear model components outperforming other existing methods (including perturbation) by a large margin.

\begin{figure}
    \centering
    \includegraphics[width=0.99\linewidth]{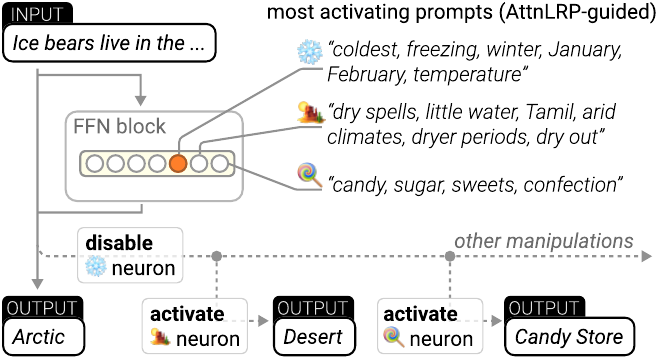}
    \caption{\gls{ours} combined with \gls{amax} allows to identify relevant neurons and gain insights into their encodings. This allows one to manipulate the latent representations and, \eg, to change the output ``Arctic'' (by disabling the corresponding neuron) to ``Desert'' or ``Candy Store'' (by activating the respective neurons). See also Section~\ref{experiments:understanding}.}
    \label{fig:manipulation}
\end{figure}

\paragraph{Contributions} 
In this work, we introduce \gls{ours}, an extension of \gls{lrp} within the Deep Taylor Decomposition framework~\cite{montavon2017explaining}, with the particular requirements necessary for attributing non-linear transformer components accurately.
\gls{ours} allows explaining transformer-based models with high faithfulness and efficiency, while also allowing attribution of latent neurons and providing insights into their role in the generation process (see Figure \ref{fig:manipulation}).

\begin{enumerate}
    \item We derive novel efficient and faithful \gls{lrp} attribution rules for non-linear attention within the Deep Taylor Decomposition framework, demonstrating their superiority over the state-of-the-art and successfully tackling the noise problem in ViTs.
    \item We illustrate how to gain insights into an \gls{llm} generation process by identifying relevant neurons and explaining their encodings.
    \item We provide an efficient and ready-to-use open source implementation of \gls{ours} for transformers. 
\end{enumerate}

\section{Related Work}
We present an overview of related work for various model-agnostic and transformer-specialized attribution methods.

\subsection{Perturbation \& Local Surrogates}

In perturbation analysis, such as occlusion-based attribution~\cite{zeiler2014visualizing} or SHAP~\cite{lundberg2017Shap}, the input features are repeatedly perturbed while the effect on the model output is measured~\cite{fong2017interpretable}. 
AtMan~\cite{deb2023atman} is specifically adapted to the transformer architecture, where tokens are not suppressed in the input space, but rather in the latent attention weights.

Interpretable local surrogates, on the other hand, replace complex black-box models with simpler linear models that locally approximate the model function being explained. Since the surrogate has low complexity, interpretability is facilitated. Prominent methods include LIME \cite{Ribeiro2016Why} and LORE \cite{guidotti2018local}.

While these approaches are model-agnostic and memory efficient, they have a high computational cost in terms of forward passes. 
Furthermore, explanations generated on surrogate models cannot explain the hidden representations of the original model. Finally, latent attributions wrt.\ the prediction must be computed for each layer separately, increasing the computational cost further.

\subsection{Attention-based}
These methods take advantage of the attention mechanism in transformer models. 
Although attention maps capture parts of the data distribution, they lack class specificity and do not provide a meaningful interpretation of the final prediction \cite{wiegreffe2019attention}.
Attention Rollout~\cite{abnar2020quantifying} attempts to address the issue by sequentially connecting attention maps of all layers. However, the resulting attributions are still not specific to individual outputs and exhibit substantial noise.
Hence, \cite{gildenblat2020} has found that reducing noise in attention rollout can be achieved by filtering out excessively strong outlier activations.

To enable class-specificity, the work of \cite{chefer2021transformer} proposed a novel rollout procedure wherein the attention's activation is mean-weighted using a combination of the gradient and LRP-inspired relevances. It is important to note that this approach yields an approximation of the mean squared relevance value, which diverges from the originally defined notion of ``relevance'' or ``importance'' of additive explanatory models such as SHAP~\cite{lundberg2017Shap} or LRP \cite{bach2015pixel}. 
Subsequent empirical observations by \cite{chefer2021generic} revealed that an omission of LRP-inspired relevances and a sole reliance on a positive mean-weighting of the attention's activation with the gradient improved the faithfulness inside cross-attention layers. Though, this approach can only attribute positively and does not consider counteracting evidence.

Attention-rollout based approaches, while offering advantages in terms of computational efficiency and conceptual simplicity, have significant drawbacks. 
Primarily, they suffer from a limited resolution in the input attribution maps, resulting in undesirable checkerboard artifacts cf.~Figure~\ref{fig:overview}. Moreover, they are unable to attribute hidden latent features beyond the softmax output. Consequently, these approaches only provide explanations for a fraction of the model, thereby compromising the fidelity and limiting the feasibility of explanations within the hidden space.

\subsection{Backpropagation-based}
Input $\times$ Gradient~\cite{simonyan2014deep} linearizes the model by utilizing the gradient.
However, this approach is vulnerable to gradient shattering \cite{balduzzi2017shattered, dombrowski2022towards}, leading to noisy attributions in deep models. Consequently, several works aim to denoise these attributions. SmoothGrad \cite{smilkov2017smoothgrad} and Integrated Gradients \cite{sundararajan2017axiomatic} have attempted to address this issue but have been unsuccessful in the case of large transformers, as demonstrated in the experiments in Section~\ref{experiments:evaluation}.
\cite{chefer2021transformer} adapted Grad-CAM~\cite{selvaraju2017grad} to transformer models by weighting the last attention map with the gradient.

Modified backpropagation methods, such as LRP \cite{bach2015pixel}, decompose individual layer functions instead of linearizing the entire model. They modify the gradient to produce more reliable attributions~\cite{ArrINF22}. The work~\cite{ding2017visualizing} was the first to apply standard LRP on non-linear attention layers, while \cite{voita2020analyzing} proposed an improved variant building upon the Deep Taylor Decomposition framework. Nonetheless, both variants can lead to numerical instabilities in attributing the softmax function and do not fulfill the conservation property~(\ref{lrp:conservation}) in matrix multiplication. \cite{ali2022xai} considerably improved attributions by recognizing that standard LRP rules were not suitable for these operations and proposed to exclude softmax and normalization operations from the computational graph by stopping the relevance (gradient) flow through them. However, it does not resolve the fundamental challenge of optimally applying LRP to non-linear operations.
In Appendix \ref{app:difference}, we provide a comprehensive analysis about different LRP-variants.

\section{Attention-Aware LRP for Transformers}
First, we motivate LRP in the framework of additive explanatory models. Then, we generalize the design of new rules for non-linear operations. Finally, we apply our methodology successively on each operation utilized in a transformer model to derive efficient and faithful rules.

\subsection{Layer-wise Relevance Propagation}
\label{methods:lrp}

Layer-wise Relevance Propagation (LRP) \cite{bach2015pixel, montavon2019layer} belongs to the family of additive explanatory models, which includes the well-known Shapley \cite{lundberg2017Shap}, Gradient $\times$ Input~\cite{simonyan2014deep} and DeepLIFT~\cite{shrikumar2017learning} methods.

The underlying assumption of such models is that a function $f_j$ with $N$ input features $\x = \{x_i\}_{i=1}^N$ can be decomposed into individual contributions of single input variables $R_{i\leftarrow j}$ (called ``relevances"). Here, $R_{i\leftarrow j}$ denotes the amount of output $j$ that is attributable to input $i$, which, when added together, equals (or is proportional to) the original function value. Mathematically, this can be written as:
\begin{equation}
    f_j(\x) \propto R_j = \sum_i^N R_{i\leftarrow j}
    \label{lrp:decomposition}
\end{equation}
If an input $i$ is connected to several outputs $j$, \eg, a multidimensional function \textbf{f}, the contributions of each output $j$ are losslessly aggregating together.
 \begin{equation}
    R_i = \sum_j R_{i \leftarrow j}.
    \label{eq:aggretation}
\end{equation}
This provides us with ``importance values" for the input variables, which reveal their direct contribution to the final prediction.
Unlike other methods, LRP treats a neural network as a layered directed acyclic graph,
where each neuron $j$ in layer $l$ is modeled as a function node $f^l_j$ that is individually decomposed according to Equation \eqref{lrp:decomposition}. Beginning at the model output $L$, the initial relevance value $R^L_j \propto f^L_j$ is successively distributed to its prior network neurons one layer at a time. Hence, LRP follows the flow of activations computed during the forward pass through the model in the \emph{opposite} direction, from output $f^L$ back to input layer $f^1$.

This decomposition characteristic of LRP gives rise to the important \emph{conservation property}:
\begin{equation}
\label{lrp:conservation}
     R^{l-1} = \sum_i R^{l-1}_i = \sum_{i,j} R^{(l-1, l)}_{i\leftarrow j} = \sum_j R^l_j = R^l
\end{equation}
 ensuring that the sum of all relevance values in each layer remains constant. This property allows for meaningful attribution, as the scale of each relevance value can be related to the original function output $f^L$.

\subsubsection{Decomposition through Linearization}
\label{methods:taylor}

To design a faithful attribution method, the challenge lies in identifying a meaningful distribution rule $R_{i\leftarrow j}$. Possible solutions encompass all decompositions that adhere to the conservation property \eqref{lrp:conservation}. However, for a decomposition to be considered \emph{faithful}, it should approximate the characteristics of the original function as closely as possible.

In this paper, we take advantage of the Deep Taylor Decomposition framework~\cite{montavon2017explaining} to locally linearize and decompose neural network operations into independent contributions. As a special case, we further establish the relationship between one derived rule and the Shapley Values framework in Section~\ref{methods:matrix_multiplication}.

We start by computing a first-order Taylor expansion at a reference point $\xr$. 
For the purpose of simplifying the equation, we assume that the reference point $\xr$ is constant:
\begin{align}
      f_j(\x) &= f_j(\xr) + \sum_i \textbf{J}_{ji}(\xr) \ (x_i - \xrs_i)  + \mathcal{O}(|\x -\xr|^2)
     \label{eq:taylor} \\
      &= \sum_i \textbf{J}_{ji} \ x_i + \underbrace{f_j(\xr) - \sum_i \textbf{J}_{ji}\ \xrs_i + \mathcal{O}(|\x -\xr|^2)}_{\text{bias } \tilde{b}_j} \nonumber
     \label{eq:taylor}
\end{align} 
where $\mathcal{O}$ is the approximation error in Big-$\mathcal{O}$ notation and the Jacobian $\textbf{J}$ is evaluated at reference point $\xr$, that is in the following omitted for brevity\footnote{if $\xr = \x$, this is equivalent to Gradient $\times$ Input. We have taken the DTD perspective to highlight the bias term, which is important for the upcoming discussion.}. The bias term represents the constant portion of the function and the approximation error that cannot be directly attributed to the input variables.

We substitute the layer function with its first-order expansion and assert its proportionality to a relevance value $R_j$ following Equation~\eqref{lrp:decomposition} through multiplication with a constant factor $c \in \mathbb{R}$ with $f_j(\x) \neq 0$.

\begin{equation*}
     R_j = f_j(\x) \ c  = \sum_i \underbrace{\textbf{J}_{ji} \ x_i \frac{R_j}{f_j(\x)}}_{R_{i\leftarrow j}} + \underbrace{\tilde{b}_j \frac{R_j}{f_j(\x)}}_{R_{b\leftarrow j}}
\end{equation*}
Comparing with Equation~\eqref{lrp:decomposition}, we identify $R_{i\leftarrow j}$ as the relevance assigned to the input variables and $R_{b\leftarrow j}$ as the relevance assigned to the bias term. Hence, the bias term absorbs a portion of the relevance $R_j$ that is not allocated to the input variables. This technically violates the conservation property~\eqref{lrp:conservation}, as only $R_{i\leftarrow j}$ is further distributed to prior layers reducing the amount of relevance per distribution step. However, \cite{bach2015pixel} treats bias terms as additional hidden neurons (with an activation value of one and a weight that equals the bias value, connected to the output) including them into the conservation property~\eqref{lrp:conservation}.
Consequently, we regard this relevance as preserved, rather than lost. Alternatively, to strictly enforce conservation, the absorbed relevance score of the bias term can be distributed equally among the input variables, or the bias term can be excluded completely, as explained in Appendix~\ref{app:bias}.

To obtain a propagation rule for the input variables, we apply Equation~\eqref{eq:aggretation} without the bias term. In addition, we insert a stabilizing factor $\varepsilon \ll |f_j(\x)| \in \mathbb{R}^+$ with the sign of $f_j(\x)$ to allow for the case $f_j(\x) = 0$: 
\begin{equation}
    R_i = \sum_j R_{i\leftarrow j} = \sum_j \textbf{J}_{ji} \ x_i \frac{R_j}{f_j(\x) + \varepsilon \ \text{sign}(f_j(\x))}
    \label{eq:assignment}
\end{equation}
In the following, $\text{sign}(f_j(\x)) \in \{-1, 1 \}$ is omitted for brevity. Note, that $\varepsilon$ acts as bias term and absorbs a negligible amount of the relevance.

To benefit from GPU parallelization, this formula can be written in matrix form:
\begin{equation*}  
    \Rightarrow \textbf{R}^{l-1} = \x \odot \textbf{J}^\top \cdot \textbf{R}^l \oslash (\textbf{f}(\x) + \varepsilon)
\end{equation*}
where $\odot$ denotes the Hadamard product, $\oslash$ element-wise division and $\textbf{R}^{l}$ a relevance vector at layer $l$.
This formula can be efficiently implemented in automatic differentiation libraries, such as PyTorch~\cite{paszke2019pytorch}. Compared to a basic backward pass, we have additional computational complexity for the element-wise operations.

\subsection{Attributing the Multilayer Perceptron}

Commonly a Multilayer Perceptron consists of a linear layer with a (component-wise) non-linearity producing input activations for the succeeding layer(s):
\begin{align}
    & z_j = \sum_i \textbf{W}_{ji} \ x_i + b_j \label{eq:linear} \\
    & a_j = \sigma(z_j)
\end{align}
where $\textbf{W}_{ji}$ are the weight parameters and $\sigma$ constitutes a (component-wise) non-linearity.

\subsubsection{The $\varepsilon$- and $\gamma$-LRP rule}

Linearizing linear layers \eqref{eq:linear} at any point $\x \in \mathbb{R}^N$ results in the fundamental $\varepsilon$-LRP \cite{bach2015pixel} rule
\begin{equation}
    R_i^{l-1} = \sum_j \textbf{W}_{ji} x_i \frac{R_j^l}{z_j(\x) + \varepsilon}
    \label{eq:epsilon}
\end{equation}
The bias $b_j$ of Equation~\eqref{eq:linear} and $\varepsilon$ absorb a portion of the relevance.
The proof is omitted for brevity. 
We employ the $\varepsilon$-LRP rule on all linear layers, unless specified otherwise. 

In models with many layers, the gradient of a layer can cause noisy attributions due to the gradient shattering effect \cite{balduzzi2017shattered, dombrowski2022towards}. To mitigate this noise, it is best practice to use the $\gamma$-LRP rule \cite{montavon2019layer}, an extension to improve the signal-to-noise ratio. We have observed that this effect is significantly pronounced in \glspl{vit} while LLMs lack visible noise. Therefore, we only apply the $\gamma$-LRP rule to linear layers in \glspl{vit}. For more details, please refer to Appendix~\ref{app:noise}.

\subsubsection{Handling Element-wise Non-Linearities}
\label{methods:non_linear}
Since element-wise non-linearities have only a single input and output variable, the decomposition of Equation~\eqref{lrp:decomposition} \emph{is} the operation itself. Therefore, the entire incoming relevance $R_j^{l}$ can only be assigned to the single input variable.
\begin{equation}
    R_i^{l-1} = R_i^{l}
    \label{eq:identity} 
\end{equation}
The identity rule~(\ref{eq:identity}) is applied to all element-wise operations with a single input and single output variable.

\subsection{Attributing Non-linear Attention}
\label{methods:non-linear-attention}
The heart of the transformer architecture \cite{vaswani2017attention} is non-linear attention
\begin{align}
   & \textbf{A} = \text{softmax}\left(\frac{\textbf{Q} \cdot \textbf{K}^\top}{\sqrt{d_k}}\right) \\
   & \textbf{O} = \textbf{A} \cdot \textbf{V} \label{eq:attention}\\
   & \text{softmax}_j(\x) = \frac{e^{x_j}}{\sum_k e^{x_k}} 
\end{align}
where ($\cdot$) denotes matrix multiplication, $\textbf{K} \in \mathbb{R}^{b \times s_k \times d_k}$ is the key matrix, $\textbf{Q} \in \mathbb{R}^{b \times s_q \times d_k}$ is the queries matrix, $\textbf{V} \in \mathbb{R}^{b \times s_k \times d_v}$ the values matrix, and $\textbf{O} \in \mathbb{R}^{b \times s_k \times d_v}$ is the final output of the attention mechanism.
$b$ is the batch dimension including the number of heads, and $d_k, d_v$ indicate the embedding dimensions, and $s_q, s_k$ are the number of query and key/value tokens.

First and foremost, the softmax function is highly non-linear. In addition, the matrix multiplication is bilinear, \ie, linear in both of its input variables. 
In the following, we will derive relevance propagation rules for each of these operations, taking into account considerations of efficiency.

\subsubsection{Handling the Softmax Non-Linearity}
\label{methods:softmax}
In Section~\ref{methods:taylor}, we present a generalized approach to linearization that incorporates bias terms, allowing for the absorption of a portion of the relevance.
However, \cite{ali2022xai} advocates for a strict adherence to the conservation property~\eqref{lrp:conservation} and argues that a linear decomposition of a \emph{non-linear} function should typically exclude a bias term.
While we see the virtue of this approach for operations such as RMSNorm~\cite{zhang2019root} or matrix multiplication, where $f(0)=0$, we contend that a linearization of the softmax function should inherently incorporate a bias term. This is due to the fact that even when the input is zero, the softmax function yields a value of $\frac{1}{N}$ (where $N$ represents the dimension of the inputs) which is analogous to a virtual bias term.

\textbf{Proposition 3.1} \textit{Decomposing the softmax function by a Taylor decomposition~\eqref{eq:taylor} at reference point \x yields the following relevance propagation rule:}
\begin{equation}
    R^{l-1}_i = x_i (R^{l}_i - s_i \sum_j R^{l}_j)
    \label{prop:3.1}
\end{equation}
\textit{where $s_i$ denotes the $i$-th output of the softmax function. The hidden bias term, which contains the approximation error, consequently absorbs a portion of the relevance.}

The proof can be found in Appendix \ref{app:proof:softmax}. In Appendix~\ref{app:temperature}, we explore the implications of vanishing gradients and temperature scaling on attributing the softmax function, which is important when attributing softmax outside of the attention mechanism, \eg at the classification output.
Note, that the works \cite{ding2017visualizing, voita2020analyzing, chefer2021transformer, ali2022xai} propose to handle the bias term differently to strictly enforce the conservation property~\eqref{lrp:conservation}. Most variants can lead to severe numerical instabilities as discussed in Appendix~\ref{app:bias} and seen empirically in our preliminary experiments.

\subsubsection{Handling Matrix-Multiplication}
\label{methods:matrix_multiplication}
 Since $f(0, 0) = 0$ holds, it is desirable to decompose the matrix multiplication without a bias term. To achieve this, we break down the matrix multiplication into an affine operation involving summation and a bi-linear part involving element-wise multiplication. 
\begin{equation*}
     \textbf{O}_{jp} = \sum_i \underbrace{ \textbf{A}_{ji}  \textbf{V}_{ip} }_{\text{bi-linear part}}
\end{equation*}
The summation already provides a decomposition in the form of Equation~\eqref{lrp:decomposition}, and we only need to decompose the individual summands $\textbf{A}_{ji}  \textbf{V}_{ip}$.

\textbf{Proposition 3.2} \textit{Decomposing element-wise multiplication with $N$ input variables of the form
\begin{equation*}
    f_j(\x) = \prod_i^N x_{ji}
\end{equation*}
by Shapley (with baseline zero) or Taylor decomposition~\eqref{eq:taylor} at reference point \x (without bias or distributing the bias uniformely) yields the following uniform relevance propagation rule:}
\begin{equation}
    R_{ji}^{l-1} = \frac{1}{N} R_j^l\,.
    \label{eq:uniform}
\end{equation}
The proof can be found in Appendix~\ref{app:proof:multiplication}.
Consequently, the combined rule can be effectively computed using: 

\textbf{Proposition 3.3} \textit{Decomposing matrix multiplication with a sequential application of the $\varepsilon$-rule~(\ref{eq:epsilon}) and the uniform rule~(\ref{eq:uniform}) on the summands yields the following relevance propagation rule for $\textbf{A}_{ji}$:}
\begin{align}
    R_{ji}^{l-1} = \sum_p \textbf{A}_{ji} \textbf{V}_{ip} \frac{R^l_{jp}}{2 \ \textbf{O}_{jp} + \varepsilon}
    \label{prop:matrix_multiplication}
\end{align}
There is no bias term absorbing relevance, whereas $\varepsilon$ absorbs a negligible quantity. For $\textbf{V}_{ip}$, we sum over the $j$ indices.
The proof can be found in Appendix~\ref{proof:putting_together}.
By employing this rule, we maintain strict adherence to the conservation property~(\ref{lrp:conservation}), as explained in Appendix~\ref{app:proof:conservation}.

\subsubsection{Handling Normalization Layers}
\label{methods:normalization}
Commonly used normalization layers in Transformers include LayerNorm~\cite{ba2016layer} and RMSNorm~\cite{zhang2019root}. These layers apply affine transformations and non-linear normalization sequentially.
\begin{align}
   & \text{LayerNorm}(\x) = \frac{x_j - \mathbb{E}[\x]}{\sqrt{\text{Var}[\x] + \varepsilon}} \gamma_j + \beta_j \\
   & \text{RMSNorm}(\x) = \frac{x_j}{\sqrt{\frac{1}{N} \sum_k x^2_k + \varepsilon}} \gamma_j
\end{align}
where $\varepsilon, \gamma_j, \beta_j \in \mathbb{R}.$
Affine transformations such as the multiplicative weighting of the output or the subtraction of the mean value are linear operations that can be attributed by the $\varepsilon$-LRP rule. Normalization, on the other hand, is non-linear and requires separate considerations. As such, we focus on the following function:
\begin{equation}
    f_j(\x) = \frac{x_j}{g(\x)}
    \label{eq:normalization}
\end{equation}
where $g(\x) = \sqrt{\text{Var}[\x] + \varepsilon}\ \ $\ or $\ g(\x) =\sqrt{\frac{1}{N} \sum_k x^2_k + \varepsilon}$.

The work~\cite{ali2022xai} demonstrates that when linearizing LayerNorm at \textbf{x}, the bias term absorbs most of the relevance equal to $\text{Var}[\x] / (\text{Var}[\x] + \varepsilon)$, effectively absorbing 99\% of the relevance with commonly used values of $\varepsilon = 10^{-6}$ and $\text{Var}[\x] = 1$. Hence, a linearization at \x is not meaningful.
As a solution, \cite{ali2022xai} proposes to regard $g(\x)$ as a constant, which transforms the normalization operation~\eqref{eq:normalization} into a (linear) element-wise operation, on which the identity rule~\eqref{eq:identity} can be applied, as discussed in Appendix~\ref{app:difference}. In the following, we prove that this heuristic can be derived from the Deep Taylor Decomposition framework.

\textbf{Proposition 3.4} \textit{Decomposing LayerNorm or RMSNorm by a Taylor decomposition~\eqref{eq:taylor} with reference point \textbf{0} (without bias or distributing the bias uniformly) yields the identity relevance propagation rule:}
\begin{equation}
    R_i^{l-1} = R_i^{l}
\end{equation}

There is no bias that absorbs relevance.
The proof is given in Appendix~\ref{app:proof:normalization}.
This rule enforces a strict notion of conservation, while being highly efficient by excluding normalization operations from the computational graph. Experiments in Section~\ref{experiments:evaluation} provide evidence that this simplification is faithful.

\subsection{Understanding Latent Features}
\begin{figure}[t]
    \centering
    \includegraphics[width=0.99\linewidth]{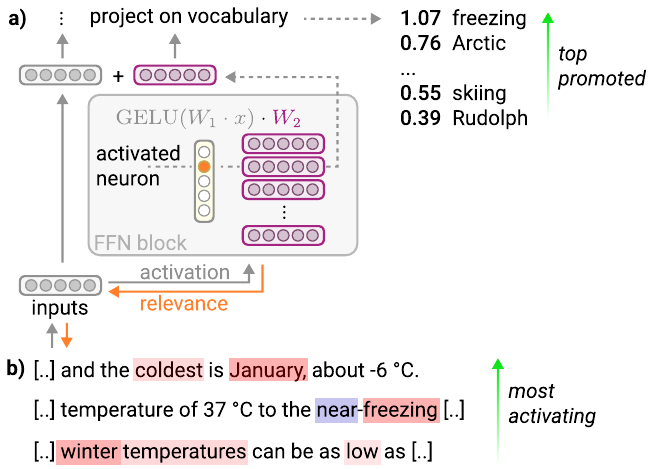}
    \caption{There are two approaches for understanding knowledge neurons: (a) Neuron \texttt{3948} at the last non-linearity in \gls{ffn} 17 of the Phi-1.5 model selects a weight row to add to the residual stream. This weight row projected on the vocabulary spans topics about ice, cold places and winter sport.
    (b) Sentences that maximally activate this neuron contain references about coldness. Attributing the neuron with \gls{ours} highlights the most relevant tokens inside the input sentences.  
    Inspired by \cite{voita2023neurons}.}
    \label{fig:latent_llm}
\end{figure}
As we iterate through each layer during the attribution process with AttnLRP, we obtain relevance values for each latent neuron as a by-product. Ranking this latent relevance enables us to identify neurons and layers that are most influential for the reasoning process of the model \cite{achtibat2023attribution}. The subsequent step is to reveal the concept that is represented by each neuron by finding the most representative reference samples that explain the neuron's encoding. A common technique is Activation Maximization (ActMax)~\cite{nguyen2016synthesizing}, where input samples are sought that give rise to the highest activation value.
We follow up on these observations and present the following strategy for understanding latent features:
(1) Collect prompts that lead to the highest activation of a unit.
(2) Explain the unit's activation using \gls{ours}, allowing to narrow down the relevant input tokens for the chosen unit.

In this work, we concentrate on knowledge neurons~\cite{dai2022knowledge, voita2023neurons} that are situated at the last non-linearity in \gls{ffn} layers $\textbf{z} = \text{GELU}(\textbf{W}_1 \x)$. These neurons possess intriguing properties, as shown in Figure~\ref{fig:latent_llm}: They encode factual knowledge and upon activation, the corresponding row of the second weight matrix $\textbf{W}_2$ is added to the residual stream directly influencing the output distribution of the model. By projecting this weight row onto the vocabulary, a distribution of the most probable tokens across the vocabulary is obtained~\cite{geva2022transformer}.
Applying \gls{ours} on \gls{amax} reference samples and projecting the weight row on the vocabulary allow us to understand in which context a neuron activates and how its activation influences the prediction of the next token. In contrast to \cite{ali2022xai}, \gls{ours} also allows analyzing the key and value linear layers inside attention modules.

\section{Experiments}

\begin{table*}[h!]
    \centering
    \caption{Faithfulness scores as area between the least and most relevant order perturbation curves \cite{blucher2024decoupling} on different models and datasets. To assess plausibility, the (top-1) accuracy along with the IoU in parentheses are depicted for SQuAD\,v2. Methods marked with $(\ast)$ have been proposed here. Additional results for ViT-L-16 and ViT-L-32 are in Appendix Table~\ref{tab:app:vit_more}.}
    \label{tab:faithfulness}    
    \begin{tabular}{@{}l@{\hspace{0.3em}}c@{\hspace{0.3em}}c@{\hspace{0.3em}}c@{\hspace{0.3em}}c@{\hspace{0.7em}}c@{}}
    
        \toprule
        \textbf{Methods}  & \textbf{ViT-B-16} & \multicolumn{2}{c}{\textbf{LLaMa 2-7b}} & \textbf{Mixtral 8x7b} & \textbf{Flan-T5-XL}\\
         & ImageNet $\uparrow$ & IMDB $\uparrow$ & Wikipedia $\uparrow$ & SQuAD\,v2 $\uparrow$ & SQuAD\,v2 $\uparrow$\\
        \midrule

        Random & \phantom{-}\phantom{-}$0.01{\,\pm\, 0.01}$ & $-0.01\tiny{\,\pm\, 0.05}$ & $-0.07\tiny{\,\pm\, 0.13}$ & \phantom{-}\phantom{-}$0.03\,\,(0.09)$ & \phantom{-}\phantom{-}$0.03\,\,(0.08)$ \\
        \midrule
        
        Input$\times$Grad~\cite{simonyan2014deep} & \phantom{-}\phantom{-}$0.80\tiny{\,\pm\, 0.03}$ & \phantom{-}\phantom{-}$0.12\tiny{\,\pm\, 0.05}$ & \phantom{-}\phantom{-}$0.18\tiny{\,\pm\, 0.13}$ & \phantom{-}\phantom{-}$0.56\,\,(0.35)$ & \phantom{-}\phantom{-}$0.60\,\,(0.39)$ \\

        IG~\cite{sundararajan2017axiomatic} & \phantom{-}\phantom{-}$1.54\tiny{\,\pm\, 0.03}$ & \phantom{-}\phantom{-}$1.23\tiny{\,\pm\, 0.05}$ & \phantom{-}\phantom{-}$4.05\tiny{\,\pm\, 0.13}$ & \phantom{-}\phantom{-}$0.68\,\,(0.44)$ & \phantom{-}\phantom{-}$0.10\,\,(0.16)$\\

        SmoothGrad~\cite{smilkov2017smoothgrad} & $-0.04\tiny{\,\pm\, 0.03}$ & \phantom{-}\phantom{-}$0.25\tiny{\,\pm\, 0.05}$ & $-2.22\tiny{\,\pm\, 0.14}$ & \phantom{-}\phantom{-}$0.47\,\,(0.24)$ & \phantom{-}\phantom{-}$0.05\,\,(0.09)$ \\
        \midrule
        
        GradCAM~\cite{chefer2021transformer} & \phantom{-}\phantom{-}$0.27\tiny{\,\pm\, 0.04}$ & $-0.82\tiny{\,\pm\, 0.05}$ & \phantom{-}\phantom{-}$2.01\tiny{\,\pm\, 0.15}$ & \phantom{-}\phantom{-}$0.82\,\,(0.72)$ & \phantom{-}\phantom{-}$0.81\,\,(0.70)$ \\%

        AttnRoll~\cite{abnar2020quantifying} & \phantom{-}\phantom{-}$1.31\tiny{\,\pm\, 0.03}$ & $-0.64\tiny{\,\pm\, 0.05}$ & $-3.49\tiny{\,\pm\, 0.15}$ & \phantom{-}\phantom{-}$0.05\,\,(0.10)$ & \phantom{-}\phantom{-}$0.02\,\,(0.08)$ \\

        Grad$\times$AttnRoll~\cite{chefer2021generic} & \phantom{-}\phantom{-}$2.60\tiny{\,\pm\, 0.03}$ & \phantom{-}\phantom{-}$1.61\tiny{\,\pm\, 0.05}$ & \phantom{-}\phantom{-}$9.79\tiny{\,\pm\, 0.14}$ & \phantom{-}\phantom{-}$0.91\,\,(0.40)$ & \phantom{-}\phantom{-}$\mathbf{0.94}\,\,(0.53)$ \\

        \midrule
        AtMan~\cite{deb2023atman} & \phantom{-}\phantom{-}$0.70\tiny{\,\pm\, 0.02}$ & $-0.20\tiny{\,\pm\, 0.05}$ & \phantom{-}\phantom{-}$3.31\tiny{\,\pm\, 0.15}$ & \phantom{-}\phantom{-}$0.86\,\,(\mathbf{0.83})$ & \phantom{-}\phantom{-}$0.88\,\,(0.80)$\\ 

        KernelSHAP~\cite{lundberg2017Shap} & \phantom{-}\phantom{-}$4.71\tiny{\,\pm\, 0.03}$ & - & - & - & - \\
        \midrule

        CP-LRP ($\varepsilon$-rule, \citet{ali2022xai}) & \phantom{-}\phantom{-}$2.53\tiny{\,\pm\, 0.02}$ & \phantom{-}\phantom{-}$1.72\tiny{\,\pm\, 0.04}$ & \phantom{-}\phantom{-}$7.85\tiny{\,\pm\, 0.12}$ & \phantom{-}\phantom{-}$0.50\,\,(0.40)$ & \phantom{-}\phantom{-}$0.91\,\,(0.83)$\\

        CP-LRP ($\gamma$-rule for ViT, as proposed here)* & \phantom{-}\phantom{-}$6.06\tiny{\,\pm\, 0.02}$ & \phantom{-}\phantom{-}- & \phantom{-}\phantom{-}- & \phantom{-}\phantom{-}- & -\\

        \gls{ours} (ours)* & \phantom{-}\phantom{-}$\mathbf{6.19}\tiny{\,\pm\, 0.02}$ & \phantom{-}\phantom{-}$\mathbf{2.50}\tiny{\,\pm\, 0.05}$ & \phantom{-}\phantom{-}$\mathbf{10.93}\tiny{\,\pm\, 0.13}$ & \phantom{-}\phantom{-}$\mathbf{0.96}\,\,(0.72)$ & \phantom{-}\phantom{-}$\mathbf{0.94}\,\,(\mathbf{0.84})$\\
        
        \bottomrule
    \end{tabular}
\end{table*}

Our experiments aim to answer the following questions:
\begin{itemize}
    \item[\textbf{(Q1)}] How faithful are our explanations compared to other state-of-the-art approaches?
    \item[\textbf{(Q2)}] How efficient is LRP compared to perturbation-based methods?
    \item[\textbf{(Q3)}] Can we understand latent representations and interact with LLMs?
\end{itemize}

\subsection{Evaluating Explanations (Q1)}
\label{experiments:evaluation}

A reliable measure of faithfulness of an explanation are input perturbation experiments~\cite{samek2016evaluating, hedstrom2023quantus}. 
This approach iteratively substitutes the most important tokens in the input domain with a baseline value. If the attribution method accurately identified the most important tokens, the model's confidence in the predicted output should rapidly decrease. The other way around, perturbing the least relevant tokens first, should not affect the model's prediction and result in a slow decline of the model's confidence.
For more details, see Appendix~\ref{sec:input_perturbation}.
Despite its drawbacks, such as potentially introducing out-of-distribution manipulations \cite{chang2018explaining} and sensitivity towards the chosen baseline value, this approach is widely adopted in the community. \cite{brocki2023feature, blucher2024decoupling} have addressed this criticism and introduced an enhanced metric by quantifying the area between the least and most relevant order perturbation curves to obtain a robust measure. Hence, we will employ this improved metric to measure faithfulness. Appendix Figure~\ref{app:fig:auc_comparison} illustrates a typical perturbation curve.

In order to assess plausibility, we utilize the SQuAD\,v2 Question-Answering (QA) dataset~\cite{rajpurkar2018know}, which includes a ground truth mask indicating the correct answer within the question. We calculate attributions for accurately answered questions and determine the top-1 accuracy of the most relevant token and the \gls{iou} between the positive attribution values and the ground truth mask. This approach assumes that the model solely relies on the information provided in the ground truth mask, which is not entirely accurate but sufficient for identifying a trend.

\subsubsection{Baselines}
We evaluate the faithfulness on two self-attention models, a ViT-B-16~\cite{DosovitskiyB0WZ21} on ImageNet~\cite{deng2009imagenet} classification and the LLaMa 2-7b~\cite{touvron2023llama} model on IMDB movie review~\cite{maas2011learning} classification as well as next word prediction of Wikipedia~\cite{wikimedia2024le}. Additional results for ViT-L-16 and ViT-L-32 are in Appendix Table~\ref{tab:app:vit_more}. To assess plausability, we employ two instruction-finetuned models on the SQuAD\,v2 dataset: the \gls{moe} model Mixtral 8x7b~\cite{jiang2024mixtral} and the encoder-decoder model Flan T5-XL~\cite{chung2022scaling}.
We denote our method as \gls{ours} and compare it against a broad spectrum of methods including Input$\times$Gradient (I$\times$G), Integrated Gradients (IG), SmoothGrad (SmoothG), Attention Rollout (AttnRoll), Gradient-weighted Attention Rollout (G$\times$AttnRoll) and Conservative Propagation (CP)-LRP. As explained in Appendix \ref{app:noise}, we propose to apply the $\gamma$-rule for \gls{ours} in the case of \glspl{vit}. For better comparison, we also included an enhanced CP-LRP baseline, which also uses the $\gamma$-rule in the \glspl{vit} experiment. The LRP variants introduced by \cite{voita2020analyzing, chefer2021transformer} are excluded due to numerical instabilities observed in preliminary experiments, see also Appendix~\ref{app:bias}.
Further, we utilize the Grad-CAM adaptation described in \cite{chefer2021transformer}. Specifically, we weight the last attention map with the gradient. For a fair comparison, we attribute all methods without the softmax at the classification output, except AtMan which relies on it.
KernelSHAP is only evaluated on vision transformers due to prohibitive computational costs on larger LLMs.
Finally, we expand upon AtMan by incorporating it into encoder-decoder models by suppressing tokens in all self-attention layers within the encoder, while only doing so in cross-attention layers within the decoder.
For AtMan, SmoothGrad and Rollout-methods we perform a hyperparameter sweep over a subset of the dataset. More details about baseline methods and the hyperparameter search are in Appendix \ref{app:details:baselines} and \ref{app:sweepsearch_sg_atman}. We illustrate example heatmaps for SQuAD\,v2 in Appendix~\ref{app:compare_squad}.

\subsubsection{Discussion}
In Table \ref{tab:faithfulness}, we can observe that \gls{ours} consistently outperforms all the state-of-the-art methods in terms of faithfulness. 
In models with a higher number of non-linearities (higher complexity), AttnLRP demonstrates substantially higher accuracy compared to CP-LRP. While the relative improvement to CP-LRP is 3\% for Flan-T5-XL, which only utilizes standard attention layers, AttnLRP achieves a remarkable 46\% improvement over CP-LRP in terms of top-1 accuracy in Mixtral 8x7b, that incorporates additional expert layers with softmax non-linearities and FFN layers with non-linear weighting. In Appendix \ref{app:ablations}, we discuss the architectural differences and conduct an ablation study on different model components to demonstrate this effect.
We also observe that gradient-based approaches significantly suffer from noisy attributions, as reflected by the low faithfulness and illustrated in example heatmaps in Appendix~\ref{app:compare_squad}. CP-LRP with $\varepsilon$ applied on all layers (as proposed in \citet{ali2022xai}), also suffers from noisy gradients in \glspl{vit}. Applying instead the $\gamma$-rule for CP-LRP and \gls{ours} in \glspl{vit} improves the faithfulness substantially.
Whereas AtMan and GradCAM do not perform well in unstructured tasks, \ie, next word prediction or classification, they achieve a high score in QA tasks. While G$\times$AttnRoll better reflects the model behavior compared to AtMan and GradCAM, it is affected by considerable background noise, resulting in a low \gls{iou} score in the SQuAD\,v2 dataset. 

\subsection{Computational Complexity and Memory Consumption (Q2)}
Table~\ref{table:efficiency} illustrates the computational complexity and memory consumption of a single LRP-based attribution and linear-time perturbation, such as AtMan or a Shapley-based method~\cite{fatima2008linear}.
Linear-time perturbation requires $N_T$ forward passes, but has only a memory requirement of $\mathcal{O}(1)$. Since LRP is a backpropagation-based method, gradient checkpointing~\cite{chen2016training} techniques can be applied. In checkpointing, LRP requires two forward and one backward pass, while the memory requirement scales logarithmic with the number of layers. In Appendix~\ref{app:benchmarking}, we benchmark energy, time and memory consumption of LRP against perturbation-based methods across context- and model-sizes.
\begin{table}[t!]
    \centering
    \caption{Computational and memory complexity of LRP-based and linear-time perturbation methods measured \wrt a single forward pass. $N_L$: number of layers, $N_T$: number of tokens}
    \label{table:efficiency}
    \begin{tabular}{lccc}
        \toprule
        \textbf{Methods}  & \textbf{Computational} & \textbf{Memory}\\
        & \textbf{Complexity} & \textbf{Consumption} \\
        \midrule
        LRP Checkpointing & $\mathcal{O}(1)$ & $\mathcal{O}(\sqrt{N_L})$\\ 
        Perturbation (linear) & $\mathcal{O}(N_T)$ & $\mathcal{O}(1)$  \\ 
        \bottomrule
    \end{tabular}
\end{table}

\subsection{Understanding \& Manipulating Neurons (Q3)}
\label{experiments:understanding}
In our investigation, we use the Phi-1.5 model \cite{li2023textbooks}, which has a transformer-based architecture with a next-word prediction objective. 
We obtain reference samples for each knowledge neuron by collecting the most activating sentences over the Wikipedia summary dataset~\cite{scheepers2017compositionality}.


To illustrate, we consider the prompt: `\texttt{The ice bear lives in the}' which gives the corresponding prediction: `\texttt{Arctic}'. Using \gls{ours}, we determine the most relevant layers for predicting `\texttt{Arctic}' as well as the specific neurons within the \gls{ffn} layers contributing to this prediction.
Our analysis reveals that the most relevant neurons after the first three layers are predominantly situated within the middle layers. Notably, one standout neuron \texttt{\#3948} in \texttt{layer 17} activates on reference samples about cold temperatures, as depicted in Figure~\ref{fig:latent_llm}.
This observation is further validated by projecting the weight matrix of the second FFN layer onto the vocabulary. The neuron shifts the output distribution of the model to cold places, winter sports and animals living in cold regions. 

Analogously, for the prompt `\texttt{Children love to eat sugar and}' with the prediction `\texttt{sweets}', the most relevant neuron's (\texttt{layer 18}, neuron \texttt{\#5687}) projection onto the vocabulary signifies a shift in the model's focus towards the concept of candy, temptation and sweetness in the vocabulary space. 
We interact with the model by deactivating neuron \texttt{\#3948}, and strongly amplifying the activation of neuron \texttt{\#5687} in the forward pass. This manipulation yields the following prediction change:

Prompt: \texttt{Ice bears live in the}\\
Prediction:\,\,\texttt{sweet, sugary treats of the candy store.}

We further notice that neuron \texttt{\#4104} in \texttt{layer 17} encodes for dryness, thirst and sand. Increasing its activation changes the output to `\texttt{desert}' (illustrated in Figure~\ref{fig:manipulation}).

With \gls{ours}, we are able to trace the most important neurons in models with billions of parameters. This allows us to systematically navigate the latent space to enable targeted modifications to reduce the impact of certain concepts (for example, `\texttt{coldness}') and enhance the presence of other concepts (for example, `\texttt{dryness}'), resulting in discernible output changes. Such an approach holds significant implications for transformer-based models, which have been difficult to manipulate and explain due to inherent opacity and size.

\section{Conclusion}
We have extended the Layer-wise Relevance Propagation framework to non-linear attention, proposing novel rules for the softmax and matrix-multiplication step and providing interpretations in terms of Deep Taylor Decomposition.
Our \gls{ours} method stands out due to its unique combination of simplicity, faithfulness, and efficiency. 
We demonstrate its applicability both for LLMs as well as ViTs, utilizing the denoising effect of the $\gamma$-rule.
In contrast to other backpropagation-based approaches, \gls{ours} enables the accurate attribution of neurons in latent space (also within the attention module), thereby introducing novel possibilities for real-time model interaction and interpretation.

\section*{Limitations \& Open Problems}
Adjusting the $\gamma$-parameter in ViTs remains crucial to achieve accurate attributions.
To reduce memory consumption, the impact of quantization on attributions and custom GPU kernels for LRP rules should be investigated. 

\section*{Acknowledgements}
We extend our heartfelt gratitude to Leila Arras for her invaluable feedback and to Johanna Vielhaben for improving our faithfulness metric. We also thank Patrick Kahardipraja, Daniel Becking and Maximilian Ernst for their insightful comments.

\section*{Impact Statement}
This work establishes the foundations that make it possible to systematically analyze and debug transformer-based AI systems, thereby minimizing the occurrence of false or misleading outputs (hallucination) and mitigating biases that may arise from training data or algorithmic processes. Particularly, it opens up the door for future applications of transformer-based AI systems in critical domains such as healthcare and finance, where the ability to explain the model behavior is often a (legal) requirement. The high computational efficiency of our method significantly reduces the energy usage and consequently also the financial overhead and environmental impact associated with the explanation, which will result in a broader adoption of XAI for transformers.

\newpage
\bibliographystyle{apalike} 
\bibliography{bibliography}

\newpage
\appendix
\renewcommand{\thetable}{\thesection.\arabic{table}}
\renewcommand{\thefigure}{\thesection.\arabic{figure}}

\section*{Appendix}
\section{Appendix I: Methodological Details}
This appendix provides further details on the methods presented in the paper. In particular, we focus on the \gls{ours} method, provide implementation details, discuss the stability of the bias term, highlight the difference between other LRP variants, discuss the noise problem in Vision Transformers and illustrate the effects of temperature scaling on attributing the softmax function. Finally, we provide proofs for the four propositions presented in the main paper.

\subsection{Details on Baseline Methods}
\label{app:details:baselines}
In the following, we present an overview of the baseline methods and their hyperparamter choices.

\subsubsection{Input $\times$ Gradient}\label{app:sec:ixg}
Gradients are one of the most straightforward approaches to depict how sensitive the trained model is with respect to each individual given feature (traditionally of the input space). 
By weighting the gradient with the input features, the model is locally linearized \cite{simonyan2014deep}:
\begin{equation}
    \centering
    \text{I}\times\text{G(\textbf{x})} = \frac{\partial f_{c}(\x)}{\partial \x} \times \x
    \label{eq:inputtimesgrad}
\end{equation}

Due to the gradient shattering effect \cite{balduzzi2017shattered} which is a known phenomenon (especially in the ReLU-based CNNs), heatmaps generated by I$\times$G are very noisy, making them in many cases not meaningful.

\subsubsection{Integrated Gradients}
To tackle the noisiness of I$\times$G, the idea to integrate gradients along a trajectory has been proposed. Here, the gradients of different ($m$) interpolated versions of the input $\x$, noted by $\x^{\prime}$, are integrated as \cite{sundararajan2017axiomatic}:
\begin{align}
    \begin{split}
    \text{IG}(\x) &= (\x - \x^{\prime}) \int_{\alpha = 0}^{1} \frac{\partial f_{j}(\x^{\prime} + \alpha \times (\x - \x^{\prime}))}{\partial \x} d\x \\
    &\approx (\x - \x^{\prime}) \sum_{k=1}^{m} \frac{\partial f_{j}(\x^{\prime} + \frac{k}{m} \times (\x - \x^{\prime}))}{\partial \x} \times \frac{1}{m}
    \end{split}\label{eq:integratedgrad}
\end{align}
We utilize \texttt{zennit}~\cite{anders2021software} and its default settings to compute Integrated Gradients attribution maps \ie $m=20$.

\subsubsection{SmoothGrad}
A different technique towards the reduction of noisy gradients is smoothing the gradients \cite{smilkov2017smoothgrad} through generating ($m$) various samples in the neighborhood of input $\x$ as $\x_{\varepsilon} = \x + \mathcal{N}(\mu,\,\sigma^{2})$ and computing the average of all gradients:
\begin{equation}
    \centering
    \text{SmoothGrad}(\x) = \frac{1}{m} \sum_{1}^{m} \frac{\partial f_{j}(\x_{\varepsilon})}{\partial \x_{\varepsilon}}
\end{equation}
In this work, we set $\mu=0$ and perform a hyperparameter search for $\sigma$ to find the optimal parameter. We utilize \texttt{zennit}~\cite{anders2021software} and its default settings to compute SmoothGrad attribution maps \ie $m=20$.

\subsubsection{Attention Rollout}
Self-Attention rollout~\cite{abnar2020quantifying} capitalizes on the intrinsic nature of the attention weights matrix $\textbf{A} \in \mathbb{R}^{b\times s_q\times s_k}$ as a representative measure of token importance. It generates a $s_q\times s_k$ matrix where each row is normalized to form a probability distribution, representing the importance of each query token to all key tokens. The attention scores along the head dimension are averaged:
\begin{equation*}
    \Bar{\textbf{A}} = \mathbb{E}_b[\textbf{A}]
\end{equation*}
where $\mathbb{E}_b$ denotes the expectation along the head dimension $b$ of the attention map. 
To compute the relevance of hidden layer tokens ($h$) to the original input tokens ($i$), an iterative multiplication of the attention matrices on the left side is sufficient. Hence, the key dimension represents the inputs and the query dimension the outputs. To account for the residual connection through which the information of the previous tokens flows, an identity matrix \textbf{I} is added:
\begin{equation}
    \textbf{R}^{h, i}_{k} = (\textbf{I} + \Bar{\textbf{A}}^{h, h}) \cdot \textbf{R}^{h, i}_{k-1}
    \label{app:eq:rollout}
\end{equation}
where $k=1$ corresponds to the input layer and $\textbf{R}^{h, i}_{0}$ is initialized with the identity matrix \textbf{I}, $h$ denotes the hidden feature space, and $i$ stands for input dimension.

\cite{chefer2021generic} build upon self-attention rollout and weights the attention matrix with the gradient. Additionally, the weighted attention map is denoised by computing the mean value of only positive values.
\begin{equation*}
    \Bar{\textbf{A}} = \mathbb{E}_b[(\nabla \textbf{A} \odot \textbf{A})^+] 
\end{equation*}
For encoder-decoder models, \cite{chefer2021generic} present several additional considerations that are not mentioned here for brevity.

\cite{gildenblat2020} notes, that the rollout attributions can further be improved by discarding outlier values. For that, we define a discard threshold $dt \in [0, 1]$ used to compute the quantile \(Q(dt)\), where $dt$ represents the proportion of the data below the quantile \eg with cumulative distribution function $P(\Bar{\textbf{A}} \leq Q(dt)) = dt$.
\[ \Bar{\textbf{A}}_{m,n} = \begin{cases}  0 & \text{if } \Bar{\textbf{A}}_{m,n} > Q(dt) \\ \Bar{\textbf{A}}_{m,n} & \text{otherwise} \end{cases} \]

\subsubsection{AtMan}
AtMan~\cite{deb2023atman} perturbs the pre-softmax activations along the $k$-dimension:
\begin{align*}
    & \textbf{H} = \textbf{Q} \cdot \textbf{K}^\top\\
    & \tilde{\textbf{H}} = \textbf{H} \odot (\textbf{1}-\textbf{p}^i )
\end{align*}
where $\textbf{H} \in \mathbb{R}^{b\times s_q\times s_k}$, and $\textbf{1} \in [1]^{b\times s_q\times s_k}$ a matrix containing only 1. $\textbf{p}^i$ denotes a matrix $\in \mathbb{R}^{b\times s_q\times s_k}$ with 
\begin{equation*}
   \textbf{p}_{lmn}^i = 
   \begin{cases} 
        p \text{ for } n=i \\
        0 \text{ for } n \neq i
    \end{cases}
\end{equation*}
Thus, for a single token $i \in \{1, 2, ..., N\}$, we suppress all values along the column/key-dimension with a suppression factor $p$.
The suppression factor is a hyperparameter that must be tuned to the dataset and model.
For ViTs, additional cosine similarities are computed to suppress correlated tokens as detailed in \cite{deb2023atman}. For that, an additional hyperparameter denoted as $t$ for threshold must be optimized in ViTs only.

\subsubsection{KernelSHAP}
LIME computes attributions by fitting an additive surrogate model \cite{Ribeiro2016Why}. KernelSHAP~\cite{lundberg2017Shap} is a special case of LIME, that sets the loss function, weighting kernel and regularization terms of LIME such that LIME recovers Shapley values. Hence, KernelSHAP allows theoretically to obtain Shapley Values more efficiently than directly computing Shapley Values.

To apply KernelSHAP in the vision domain, we divide the input image into $N$ super-pixels using the Simple Linear Iterative Clustering (SLIC) algorithm~\cite{achanta2012slic}. We use \texttt{captum}~\cite{kokhlikyan2020captum} with its default settings to compute the attributions \ie number of samples per attributions set to 2000 and baseline value set to 0.5. A baseline value of 0 resulted in lower faithfulness.
For SLIC, we set $N=100$ with compactness set to 10.


\subsection{Details on \gls{ours}}
This section provides more details on \gls{ours} and justifies the specific parameter choices made in our work (\eg, use of $\gamma$-LRP in Vision Transformers).
\subsubsection{Conservation \& Numerical Stability of Bias Terms}
\label{app:bias}
The total relevance $R_j$ of a layer output, \ie linearized function $f_j(\x) = \sum_i \textbf{J}_{ji} \ x_i + \tilde{b}_j$, is computed by summing the contributions of the input variables $R_{i\leftarrow j}$, represented by $\textbf{J}_{ji} \ x_i$, and adding the contribution of the bias term $R_{b\leftarrow j}$, represented by $\tilde{b}_j$.
\[ R_j = \sum_i R_{i\leftarrow j} + R_{b\leftarrow j}, \] 
The relevance of the input variables is solely determined by the input variables themselves
\[ R_i = \sum_j R_{i\leftarrow j} = \sum_j \textbf{J}_{ji} \ x_i \frac{R_j}{f_j(\x)}, \]
while the relevance of the bias term itself is calculated as  \[R_{b\leftarrow j} = \tilde{b}_j \frac{R_j}{f_j(\x)}. \] 

If we want to compute the relevance of the input variables $R_i$ while ensuring strict adherence to the conservation property~\eqref{lrp:conservation}, we must exclude the bias term, so that it does not absorb part of the relevance. 
In the literature, we find two common practices: Either distributing the bias term uniformely on the input variables~\cite{binder2016layer, voita2020analyzing} or applying the identity rule~\cite{ding2017visualizing, chefer2021transformer}. Both approaches can lead to severe numerical instabilities in specific cases that are challenging to identify. Therefore, we will dedicate some time to explain the issue in greater detail.

\textbf{Remark A.2.1} \textit{Enforcing strict conservation~\eqref{lrp:conservation} on a function, where 
\[\exists i,j \in \mathbb{N}: x_i = 0 \land f_j(\x) \neq 0,\]
by distributing the relevance of the bias term of its linearization uniformly on the input variables or applying the identity rule with $i=j$ may lead to numerical instabilities.}

\underline{Distributing the bias term}: We can distribute the relevance value of the bias term uniformly across the input variables by assuming that the bias term is part of the input variables: 
\[ R_j^l = \sum_i \tilde{R}_{i\leftarrow j}^{(l-1, l)} = \sum_i^N \left( R_{i\leftarrow j}^{(l-1, l)} + \frac{R_{b\leftarrow j}^{(l-1, l)}}{N} \right), \] 
where $N$ represents the number of input variables.
Hence,
\[ R_i^l = \sum_j \tilde{R}_{i\leftarrow j}^{(l-1, l)} = \sum_j \left( \textbf{J}_{ji} \ x_i + \frac{\tilde{b_j}}{N} \right) \frac{R_j^l}{f_j(\x)}. \]  
However, we may encounter numerical instabilities, but these effects will only become visible in the next sequential relevance propagation at the prior layer, not at this layer yet.
For example in the softmax function, we may encounter a situation where $\exists i,j \in \mathbb{N}$ with $x_i^{l-1} = 0$ but $f_j^l(\x) > 0$. 
If a non-zero relevance value from layer $l$ is assigned to $f_j^l(\x)$, then its relevance value is propagated to the input variable $x_i^{l-1}$ through the relevance message:
\[ \tilde{R}_{i\leftarrow j}^{(l-1, l)} = \frac{\tilde{b}_j}{N} \frac{R_j^l}{f_j^l(\x) + \varepsilon} \]  

Assuming we apply the $\varepsilon$-rule in succession, the relevance in the prior layer is given by:
\[ R_{k\leftarrow i}^{(l-2, l-1)} = \textbf{J}_{ik} \ x_k^{l-2} \frac{R_i^{l-1}}{ 0 +\varepsilon}\]
Here, we divide by $x_i^{l-1}=f_i^{l-1}=0$. Since $\varepsilon$ is very small, this term explodes and causes numerical instabilities. Hence, assigning a non-zero relevance value to an input that equals zero leads to numerical instabilities. Note, that these instabilities would not occur if $R_i^{l-1}=0$ \eg $R_{i\leftarrow j}^{(l-1, l)}=0$.
Functions where \( f(0)=0 \) do not encounter this issue, because zero output activations will not receive any relevance in following layers \eg $R_j^{l}=0$ using all rules described in this paper.

\underline{Applying the identity rule}:
Alternatively, we can apply the identity rule as follows:
\[ R_i^{l-1} = R_i^{l} \]
Here, numerical instabilities might also arise in the subsequent relevance propagation, not at this layer. With the same reasoning as before, the identity rule propagates a non-zero relevance value to an input variable that is zero.

\underline{Omitting the bias term}:
For the sake of completeness, we mention that omitting the bias term entirely is also an option. In this case, the relevance propagation equation is:  \[ R_i^{l-1} = \sum_j \textbf{J}_{ji} \ x_i \frac{R_j^l}{\sum_i \textbf{J}_{ji} \ x_i + \varepsilon} \]

Here, we no longer divide by the original function $f_j(\x)$, but by its linearization without the bias term. However, it is important to ensure that no sign flips occur, as $\sum_i \textbf{J}_{ji} \ x_i$ might have a different sign than $f_j(\x)$. 

\textbf{Remark A.2.2} \textit{Enforcing strict conservation~\eqref{lrp:conservation} by omitting the bias term of a linearization~\eqref{eq:taylor} can lead to sign flips in the relevance scores.}

\underline{Summary}:
In summary, applying the identity rule, distributing its relevance value uniformly across the input variables or omitting the bias term completely are possible approaches, but they have their considerations and potential challenges. Regarding the softmax non-linearity, \cite{voita2020analyzing} distributes the bias term equally on all input variables, while \cite{ding2017visualizing, chefer2021transformer} apply the element-wise identity rule~\eqref{eq:identity}. Both variants can lead to severe numerical instabilities.

\subsubsection{Highlighting the difference between various LRP methods}
\label{app:difference}
In Table~\ref{table:lrpcomparison}, we illustrate the different strategies employed for LRP in the past.

\underline{Softmax}:
\cite{voita2020analyzing} linearizes at \x but distributes the bias term equally on all input variables, while \cite{ding2017visualizing, chefer2021transformer} apply the element-wise identity rule~\eqref{eq:identity}. More specifically, \cite{ding2017visualizing} did not discuss the softmax function explicitly, but they skip all non-linear activation functions. Therefore, we assume that \cite{ding2017visualizing} applies the identity rule also to the softmax function.
Both variants enforce a strict notion of the conservation principle~\eqref{lrp:conservation}, but can lead to severe numerical instabilities as discussed in Appendix~\ref{app:bias}. \cite{ali2022xai} regards the attention matrix \textbf{A} in Equation~\eqref{eq:attention} as constant, attributing relevance solely through the value path by stopping the relevance flow through the softmax. Consequently, the query and key matrices can no longer be attributed, which reduces the faithfulness and makes latent explanations in query and key matrices infeasible.
Finally, AttnLRP linearizes at \x with a bias term that absorbs part of the relevance. The presence of a bias term in AttnLRP is justified because the softmax function yields a value of 1/N even when the input is zero. This is analogous to a bias term and is necessary to account for this behavior. This ensures not only numerically robust attributions, but also improves the faithfulness considerably.

\begin{figure*}[h!] 
  \centering
  \includegraphics[width=1\linewidth]{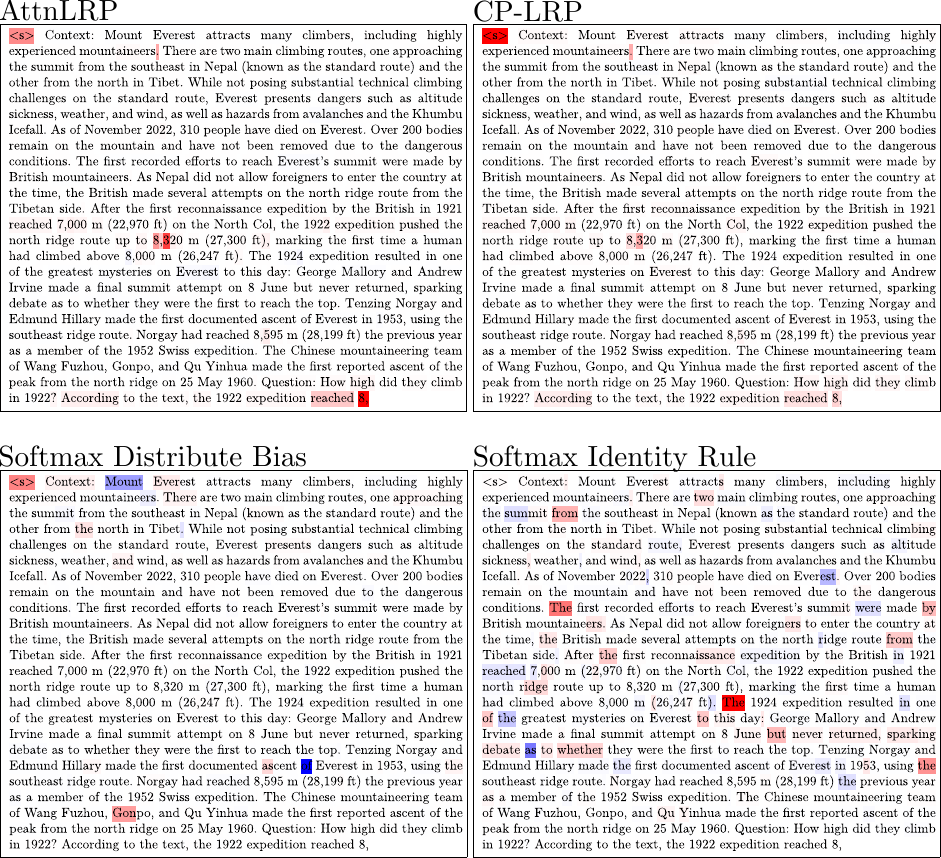}
   \caption{Comparison of four different LRP variants computed on a LLaMa 2-7b model. The given section is from the Wikipedia article on Mount Everest. The model is expected to provide the next answer token for the question \texttt{`How high did they climb in 1922? According to the text, the 1922 expedition reached 8,'}. For the correctly predicted token \texttt{3} the attribution is computed. Distributing the bias uniformely on the input variables (Softmax Distribute Bias) or applying the identity rule (Softmax Identity Rule) leads to numerical instabilities. For ``Softmax Distribute Bias" and ``Softmax Identity Rule", we applied \gls{ours} rules on all layers except for the softmax function.
   \gls{ours} highlights the correct token the strongest, while CP-LRP focuses strongly on the start-of-sequence \texttt{<s>} token and exhibits more background noise \eg irrelevant tokens such as `Context', `attracts', `Everest' are highlighted, while \gls{ours} does not highlight them or assigns negative relevance.}
\label{app:fig:compare_everest}
\end{figure*}

In Figure~\ref{app:fig:compare_everest}, we illustrate different attribution maps for all four options to handle the softmax function. The given section is from the Wikipedia article on Mount Everest. The model is expected to provide an answer for the question \texttt{`How high did they climb in 1922?'} and for the correctly predicted next token \texttt{3} of the answer \texttt{`According to the text, the 1922 expedition reached 8,'} is the attribution computed by initializing the relevance at the predicted token with its logit value.

While the relevance values for AttnLRP or CP-LRP are between $[-4, 4]$, distributing the bias uniformely on the input variables or applying the identity rule leads to an explosion of the relevances between $[-10^{15}, 10^{15}]$. As a consequence, the heatmaps resemble random noise.
\gls{ours} highlights the correct token the strongest, while CP-LRP focuses strongly on the start-of-sequence \texttt{<s>} token and exhibits more background noise \eg irrelevant tokens such as `Context', `attracts', `Everest' are highlighted, while \gls{ours} does not highlight them or assigns negative relevance.
In Appendix~\ref{app:compare_squad}, we compare also other baseline methods. Note, that the model attends to numerous tokens within the text which enables it to derive conclusions. Consequently, an attribution that reflects the model behavior will highlight more than just the single accurate answer token. The faithfulness experiments in Table~\ref{tab:faithfulness} demonstrate, that \gls{ours} captures the model reasoning most accurately. 

\underline{Matrix Multiplication}:
Applying the $\varepsilon$-rule~\eqref{eq:epsilon} on \emph{bi-linear} matrix multiplication~\eqref{eq:attention} violates the conservation property~\eqref{lrp:conservation} as proved in Appendix~\ref{app:proof:conservation}.
To the best of our knowledge, \cite{ding2017visualizing} applies the standard $\varepsilon$-rule. \cite{voita2020analyzing} utilizes the $z^+$-rule \eqref{eq:appendix:zplus-rule}, that similar to the $\varepsilon$-rule also violates the conservation property~\eqref{lrp:conservation} in bi-linear matrix multiplication (proof in Appendix~\ref{app:proof:conservation} is valid for $z^+$-rule).
While \cite{chefer2021transformer} also applies the $\varepsilon$-rule,
an additional normalization step is performed by dividing both arguments by the summation of its absolute values. This ensures conservation but is not conform with the DTD framework.
\cite{ding2017visualizing, chefer2021transformer} set the $\varepsilon$ parameter to $0$, which may increases numerical instabilities. Hence, we call their LRP variants in Table~\ref{table:lrpcomparison} $0$-LRP.

Since \cite{ali2022xai} regards the softmax output as constant and does not propagate relevance through it, the matrix multiplication is not bi-linear anymore, but becomes linear. Hence, the application of the $\varepsilon$-rule does not violate the conservation principle and attributes only the value path. \cite{ali2022xai} sets the $\varepsilon$ parameter to zero, hence we call their LRP variant in Table~\ref{table:lrpcomparison} $0$-LRP.

Finally, AttnLRP also applies the $\varepsilon$-rule on bi-linear matrix multiplication. In addition, a novel uniform rule~\eqref{eq:uniform} derived from the DTD framework is incorporated that ensures conservation and high faithfulness.

\underline{Layer Normalization}:
The works \cite{chefer2021transformer, ali2022xai} and AttnLRP apply the identity rule on normalization functions~\eqref{eq:normalization}, while using the $\varepsilon$-rule on all linear components of LayerNorm, if applicable. 
More specifically, \cite{ali2022xai} proposes to regard $g(\x)$ in \eqref{eq:normalization} as a constant, which transforms the normalization operation, and hence the complete LayerNorm layer, into a linear layer, on which the $\varepsilon$-rule is applied. However, this is similar to applying the identity rule~\eqref{eq:identity} on the normalization itself, because it becomes element-wise with a single input and output variable (see Section~\ref{methods:non_linear}), while applying the $\varepsilon$-rule on all other linear components of LayerNorm. 
\cite{voita2020analyzing} linearizes at \x and distributes the bias term equally on all input variables, which can lead to numerical instabilities as discussed in Appendix~\ref{app:bias}.

\underline{Vision Transformer}: The studies by \cite{ding2017visualizing, voita2020analyzing, ali2022xai} concentrate on the attribution in natural language processing (NLP) models and do not address vision transformers. Their methodologies, as demonstrated in Table~\ref{tab:faithfulness} (and Appendix~\ref{app:bias}), fail when implementing the $\varepsilon$-rule, leading to gradient shattering and low faithfulness. To mitigate noisy attributions, \cite{chefer2021transformer} suggests employing on-top of LRP an attention rollout~\cite{abnar2020quantifying} procedure which is additionally enhanced via gradient weighting. 
This yields an approximation of the mean squared relevance value, which diverges from the originally defined notion of ``relevance" or ``importance" of additive explanatory models. Subsequent empirical observations by \cite{chefer2021generic} revealed that an omission of LRP-inspired relevances and a sole reliance on a positive mean-weighting of the attention’s activation with the gradient improved the faithfulness. Though, this approach can only attribute positively, does not consider counteracting evidence, and does not allow to attribute latent neurons outside the attention's softmax output.
AttnLRP, in contrast, adopts the $\gamma$-rule instead of the $\varepsilon$-rule in linear layers, achieving highly faithful attributions without the necessity for a rollout mechanism. However, the $\gamma$ parameter must be tuned to the model and dataset to obtain optimal attributions, as discussed in Appendix~\ref{app:noise}.

\begin{table*}[t!]
    \centering
    \caption{Conceptual differences between various-LRP methods and their implications. ``Layer Normalization" refers here only to the normalization function~\eqref{eq:normalization} itself and not to the learnable parameters of LayerNorm or RMSNorm.}
    \begin{tabular}{lccc}
        \toprule
        \textbf{Methods}  & \textbf{Softmax} & \textbf{Matrix Multiplication}  & \textbf{Layer Normalization}\\
        \hline
        
        \cite{ding2017visualizing} & Identity rule & $0$-LRP & not available \\
        &  & (bi-linear) &  \\
        & $\Rightarrow$ unstable (Appendix \ref{app:bias}) & $\Rightarrow$ violates conservation  &  \\
        \midrule
        
        \cite{voita2020analyzing} & Taylor decomposition at \x & $z^+$-LRP & Taylor decomposition at \x \\  
         & (distributes the bias uniformly) & (bi-linear)  & (distributes the bias uniformly) \\ & $\Rightarrow$ unstable (Appendix \ref{app:bias}) & $\Rightarrow$ violates conservation & $\Rightarrow$ unstable (Appendix \ref{app:bias})\\ 
         \hline
         
        \cite{chefer2021transformer} & Identity rule & $0$-LRP  & Identity rule\\
         &  & \& post-hoc normalization & \\
         &  & (bi-linear) & \\ 
         & $\Rightarrow$ unstable (Appendix \ref{app:bias}) & $\Rightarrow$ ensures conservation  & $\Rightarrow$ ensures conservation \\
         &  & & \& faithful \\ 
         \hline
         
        \cite{ali2022xai} & Regarded as constant & $0$-LRP & Identity rule \\ 
        &  & (linear only) &  \\ &  \\ 
         & $\Rightarrow$ stable \& no attribution & $\Rightarrow$ ensures conservation & $\Rightarrow$ ensures conservation \\
         & inside attention module & & \& faithful \\ 

        \hline
        \gls{ours} & Taylor decomposition at \x  & $\varepsilon$-LRP & Identity rule \\ 
         & (with bias) & \& uniform rule &  \\ 
         &  & (bi-linear) &  \\ 
         & $\Rightarrow$ stable \& faithful 
         & $\Rightarrow$ ensures conservation &  $\Rightarrow$ ensures conservation \\
         &  &  \& faithful & \& faithful\\
        
        \bottomrule
        \label{table:lrpcomparison}
    \end{tabular}
\end{table*}

\subsubsection{Tackling Noise in Vision Transformers}
\label{app:noise}

Since backpropagation-based attributions utilize the gradient, they may produce noisy attributions in models with many layers, where gradient shattering and noisy gradients appear \cite{balduzzi2017shattered, dombrowski2022towards}.
Hence, various adaptions of the $\varepsilon$-LRP rule were developed to strengthen the signal-to-noise ratio by dampening counter-acting activations \cite{bach2015pixel, montavon2019layer}. Here, we use the generalized $\gamma$-rule that encompasses all other proposed rules in the literature \cite{montavon2019layer}.
Let $z_{ij}$ be the contribution of input $i$ to output $j$, \eg $\textbf{W}_{ji} x_i$, and $z_j$ the neuron output activation. Then depending on the sign of $z_j$:
\begin{align} \label{eq:appendix:gamma-rule-general}
    R^{(l,\: l+1)}_{i \leftarrow j} & = 
    \begin{cases}
            \frac{z_{ij} + \gamma z_{ij}^+}
    {z_j + \gamma \sum_k z_{kj}^+}
    R_j^{l+1} & \text{if } z_j>0  \\ 
            \frac{z_{ij} + \gamma z_{ij}^-}
    {z_j + \gamma \sum_k z_{kj}^-}
    R_j^{l+1} & \text{else}
    \end{cases}
\end{align}
with $\gamma \in \mathbb{R}^{>0}$, $(\cdot)^+=\text{max}(\cdot \ , 0)$ and $(\cdot)^-=\text{min}(\cdot \ , 0)$.
If $\gamma = \infty$, it is equivalent to the LRP $z^+$-rule, which is given as
\begin{align} \label{eq:appendix:zplus-rule}
R^{(l,\: l+1)}_{i \leftarrow j} & =  \frac{(w_{ij} x_i)^+}{z_j^+}R_j^{l+1}
\end{align}
by only taking into account positive contributions $z_j^+ = \sum_i (w_{ij} x_i)^+$ with $(\cdot)^+ =\max(0, \cdot)$.

Remarkably, our observations reveal that attributions in LLMs demonstrate high sparsity and lack visible noise, while \glspl{vit} are susceptible to gradient shattering. 
We hypothesize that the discrete nature of the text domain may affect robustness \cite{mao2021discrete}. Therefore, we only apply the $\gamma$-rule in \glspl{vit} in the convolutional and linear \gls{ffn} layers outside the attention module. 
To further increase the faithfulness, the $\gamma$-rule can be also applied on softmax layers. Since the output of the softmax is always greater than zero, we apply the simplified $z^+$-rule (special case of $\gamma$-rule).
The $z^+$-rule applied on a linearization~\eqref{eq:taylor} for softmax results in:

\begin{equation}
    R_i^{l-1} = \sum_j (\textbf{J}_{ji} \ x_i)^+ \frac{R_j^l}{\sum_k (\textbf{J}_{jk} \ x_k)^+ + \tilde{b}_j^+}
\end{equation}

This formula is computationally more expensive to evaluate than the original rule for softmax derived in Proposition~3.1. 
Should efficiency be a priority, it is recommended to bypass the softmax layer as proposed in CP-LRP, which prevents relevance from passing through the softmax function and reduces gradient shattering caused by this layer. Then, for all other components of the model, AttnLRP rules are recommended, with the application of the $\gamma$-rule to linear layers only. This is especially true, given that the discrepancy in faithfulness between AttnLRP and $\gamma$CP-LRP is minimal for standard vision architectures~\cite{DosovitskiyB0WZ21} evaluated in Table~\ref{tab:faithfulness} and Table~\ref{tab:app:vit_more}, that incorporate only standard attention and FFN layers.

\subsubsection{Impact of Temperature Scaling on the Softmax Rule}
\label{app:temperature}
Temperature scaling controls the entropy within the softmax probability distribution, thereby influencing the predictability of subsequent next token predictions at the classification output. A high temperature value tends to flatten the softmax output distribution (more randomness), whereas a small temperature parameter sharpens the distribution (less randomness). This scaling is done by dividing the input \x by the temperature \( T \in \mathbb{R} \) prior to applying the softmax function.

\begin{equation*}
    s_j(\x) = \frac{e^{x_j/T}}{\sum_i e^{x_i/T}}
\end{equation*}

Recall that the derivative of the softmax function has two cases, which depend on the output $j$ and input $i$ indices:
\begin{equation*}
    \frac{\partial s_j}{\partial x_i} = 
    \begin{cases}
      s_j(1-s_j) & \text{for } i=j \\
        -s_j s_i & \text{for } i \neq j
    \end{cases}
\end{equation*}

In the scenario where $s_j \approx 1$, the derivative for $i=j$ vanishes. This occurs \eg for extremely low temperature values or exceptionally high confidence in the model's classification output. This poses an issue for the Deep Taylor Decomposition \cite{montavon2017explaining} derived in Section~\ref{methods:taylor}, because DTD decomposes the softmax function by utilizing the gradient (jacobian) term $\textbf{J}_{ji} x_i$ for calculating attributions. If the gradient vanishes, the bias term will capture all the relevance, stopping the relevance flow altogether. This effect is also generally described by \cite{shrikumar2017learning}.

Within the attention mechanism, this limitation is circumvented by multiplying the softmax output with the value path, ensuring that relevance is transmitted via the uniform rule to the value path, akin to CP-LRP (refer to Appendix~\ref{app:difference}). However, in instances where the softmax function is utilized independently, this becomes problematic as the relevance flow could be distorted.

To see this, consider Proposition 3.1~\eqref{prop:3.1}:
$R^{l-1}_i$ might be zero if $R^{l}_j = 0 \ \forall j \neq i$ with $s_i = 1$:
\begin{equation*}
    R^{l-1}_i = x_i (R^{l}_i - s_i \sum_j R^{l}_j) = x_i (R^{l}_i - R^{l}_i) = 0
\end{equation*}
Therefore, we suggest utilizing an increased temperature scaling value when explaining the softmax classification output to prevent that the softmax saturates \ie the gradient vanishes. Nonetheless, the attribution of the classification output has not been investigated in this work (the softmax layer is always removed and LRP only applied to the logit outputs). An analysis of these effects remain an
interesting topic for future work.

\subsection{Proofs}
In the following, we provide proofs for the rules presented in the main paper, and that the application of the $\varepsilon$-rule on bi-linear matrix multiplication violates the conservation property.
\subsubsection{Proposition 3.1: Decomposing Softmax}
\label{app:proof:softmax}
In this subsection, we demonstrate the decomposition of the softmax function by linearizing~\eqref{eq:taylor} it at \x. We begin by considering the softmax function:
\begin{equation*}
    s_j(\x) = \frac{e^{x_j}}{\sum_i e^{x_i}}
\end{equation*}

The derivative of the softmax has two cases, which depend on the output and input indices $i$ and $j$:
\begin{equation*}
    \frac{\partial s_j}{\partial x_i} = 
    \begin{cases}
      s_j(1-s_j) & \text{for } i=j \\
        -s_j s_i & \text{for } i \neq j
    \end{cases}
\end{equation*}

Consequently, a Taylor decomposition~\eqref{eq:taylor} yields:
\begin{equation*}
    f_j(\x) = s_j \left( x_j - \sum_i s_i x_i \right) +  \tilde{b}_j
\end{equation*}
We differentiate between two cases, namely (i) when we attribute relevance from output $j$ to input $i \neq j$ and (ii) when we attribute from output $j$ to input $i=j$.
\begin{equation*}
    R_{i\leftarrow j}^{(l-1, l)} = 
    \begin{cases}
       (x_i - s_i x_i) \ R^{l}_i & \text{for } i=j \\
       -s_i x_i \ R^{l}_j & \text{for } i \neq j
    \end{cases}
\end{equation*}
Applying equation~\eqref{eq:assignment}, we obtain:
\begin{equation*}
    R^{l-1}_i = \sum_j R_{i\leftarrow j}^{(l-1, l)} = x_i (R^{l}_i - s_i \sum_j R^{l}_j)
\end{equation*}
In Appendix~\ref{app:temperature}, we discuss the implications of vanishing gradients and temperature scaling on attributing the softmax function.

\subsubsection{Proposition 3.2: Decomposing Multiplication}


\label{app:proof:multiplication}

The aim in is subsection is to decompose the multiplication of $N$ input variables.
\begin{equation*}
    f_j(\x) = \prod_i^N x_{ji}
\end{equation*}

We start by performing a Taylor decomposition~\eqref{eq:taylor}, then we derive the same decomposition with Shapley.

\underline{Taylor decomposition}: The derivative is
\begin{equation*}
    \frac{\partial f_j}{\partial x_{ji}} = \prod_{k \neq i}^N x_{jk}
\end{equation*}
Consequently, a Taylor decomposition~\eqref{eq:taylor} at $\x$ yields
\begin{equation*}
    f_j(\x) = \sum_i^N \frac{\partial f_j}{\partial x_{ji}} x_{ji} + \tilde{b}_j = N  \prod_{k}^N x_{jk} + \tilde{b}_j = N f_j(\x) + \tilde{b}_j
\end{equation*}
We can either omit the bias term or equally distribute it on the input variables to strictly enforce the conservation property~\eqref{lrp:conservation}. Here, we demonstrate how to distribute the bias term uniformly.
\begin{equation*}
    R_{ji\leftarrow j}^{(l-1, l)} = \left(f_j + \frac{\tilde{b}_j}{N}\right) \frac{R_j^l}{ N f(\x)_j + \tilde{b}_j} = \frac{1}{N} R_j^l
\end{equation*}
Since each input with index $ji$ at layer $l-1$ is only connected to one output with index $j$ at layer $l$, we have only a single relevance propagation message. Hence, it follows from Equation~\eqref{eq:aggretation}:
\begin{equation*}
    R_{ji}^{l-1} = R_{ji\leftarrow j}^{(l-1, l)} = \frac{1}{N} R_j^l
\end{equation*}

For omitting the bias term, repeat the proof with $\tilde{b}_j=0$.

\underline{Shapley}: The Shapley value \cite{lundberg2017Shap} is defined as:
\begin{equation}
     \phi_i(f) = \sum_{\substack{S \subseteq N \\ i \notin S}} \frac{|S|!(N-|S|-1)!}{N!} \left( f(S \cup \{i\}) - f(S) \right) 
\end{equation}
where 
    $\phi_i(v)$ is the Shapley value of the feature $i$ and value function
    $f$. $N$ denotes the set of all features, and $S$ denotes a feature subset (coalition). 
    
    With respect to multiplication, zero is the absorbing element. Hence, we choose zero as our baseline value, and the Shapley value function becomes:
\begin{align*}
    & f(S \cup \{i\}) = \prod_k x_k \\ 
    & f(S) = 0 \\
    & f(S \cup \{i\}) - f(S)) = \prod_k x_k
\end{align*}

The symmetry theorem \cite{fryer2021shapley} of Shapley states that the contributions of two feature values $i$ and $l$ should be the same if they contribute equally to all possible coalitions
\begin{align*}
    & f(S \cup \{i\}) = f(S \cup \{l\}) \\
    & \forall S \subseteq \{1, 2, ... N\} \backslash \{i, l\}
\end{align*}
then  $\phi_i(f) =  \phi_l(f)$.
In addition, the efficiency theorem \cite{fryer2021shapley} states that the output contribution is distributed equally amongst all features. Hence, the output contribution is equal to the sum of coalition values of all features $i$,
\begin{equation*}
    \sum_i \phi_i(f) = f(N)
\end{equation*}
Both theorems are applicable and hence it follows:
\begin{equation*}
    \phi_i(f) = \frac{1}{N} f(N)
\end{equation*}
In the case of LRP, we identify $f(N)$ as $R_j^l$ and $\phi_i(f)$ as $R_{ji}^{l-1}$.

\subsubsection{Proposition 3.3: Decomposing bi-linear Matrix Multiplication}
\label{proof:putting_together}

Consider the equation for matrix multiplication, where we treat the terms as single input variables by substituting them with $\textbf{u}_{jip}=\textbf{A}_{ji} \textbf{V}_{ip}$
\begin{equation*}
     \textbf{O}_{jp} = \sum_i  \textbf{A}_{ji} \textbf{V}_{ip} = \sum_i \textbf{u}_{jip}
\end{equation*} 
In this case, the function already is in the form of an additive decomposition~\eqref{lrp:decomposition}. Therefore,
\begin{equation*}
R_{jip\leftarrow jp}^{(l-1, l)} \propto \textbf{u}_{jip}
\end{equation*}

This can also be seen, by noticing that $\sum_i \textbf{u}_{jip}$ is a linear operation, since a single variable is left. Hence, it can be regarded as a linear layer~\eqref{eq:linear} characterized with constant weights of one and a bias of zero. We already derived the solution to applying a linearization~\eqref{eq:taylor} to a linear layer: the $\varepsilon$-rule~\eqref{eq:epsilon}. Therefore, the solution is:
\begin{equation*}
    R_{jip\leftarrow jp}^{(l-1, l)} = \textbf{u}_{jip} \frac{R_{jp}^l}{\textbf{O}_{jp} + \varepsilon} = \textbf{A}_{ji} \textbf{V}_{ip} \frac{R_{jp}^l}{\textbf{O}_{jp} + \varepsilon}
\end{equation*}

Since each input with index $jip$ at layer $l-1$ is only connected to one output with index $jp$ at layer $l$, we have only a single relevance propagation message. Hence, it follows from equation~\eqref{eq:aggretation}:
\begin{equation*}
R_{jip}^{l-1}  = R_{jip\leftarrow jp}^{(l-1, l)} 
\end{equation*}

Next, we decompose the individual terms $\textbf{u}_{jip}$ using the uniform rule from the previous Section~\ref{app:proof:multiplication} to obtain relevance messages for $\textbf{A}_{ji}$:
\begin{equation*}
    R_{ji\leftarrow jip}^{(l-1, l-1)} = \frac{1}{2} R_{jip}^{l-1}
\end{equation*}

Each input $\textbf{A}_{ji}$ is connected to $p$ outputs $\textbf{u}_{jip}$. Hence, to obtain the relevance values attributed to $\textbf{A}_{ji}$, we must aggregate all relevance messages from output ${jip}$ to inputs ${ji}$ via Equation~\eqref{eq:aggretation}:
\begin{align*}
    & R_{ji}^{l-1} = \sum_p R_{ji\leftarrow jip}^{(l-1, l-1)} = \sum_p \frac{1}{2} R_{jip}^{l-1} \\ 
    & R_{ji}^{l-1} = \sum_p \textbf{A}_{ji} \textbf{V}_{ip} \frac{R_{jp}^l}{2 (\textbf{O}_{jp} + \varepsilon)}
\end{align*}

Because $\varepsilon \ll |\textbf{O}_{jp}|$, we simplify the final solution:
\begin{equation*}
    R_{ji}^{l-1} = \sum_p \textbf{A}_{ji} \textbf{V}_{ip} \frac{R_{jp}^l}{2 \ \textbf{O}_{jp} + \varepsilon}
\end{equation*}
The proof for $\textbf{V}_{ip}$ follows a similar approach by summing over the $j$ indices instead of $p$.
In Appendix~\ref{app:proof:conservation}, we proof that this rule does not violate the conservation property~\eqref{lrp:conservation} in contrast to the standard $\varepsilon$-rule.

\subsubsection{Proposition 3.4: Layer Normalization}
\label{app:proof:normalization}

Consider layer normalization of the form
\begin{equation*}
    f_j(\x) = \frac{x_j}{g(\x)}
\end{equation*}
where $g(\x) = \sqrt{\text{Var}[\x] + \varepsilon}$ or $g(\x) =\sqrt{\frac{1}{N} \sum_k x^2_k + \varepsilon}$.
The derivative is 
\begin{equation}
    \frac{\partial f_j}{\partial x_i} = \frac{1}{g(\x)^2}
    \begin{cases}
       g(\x) - x_j \frac{\partial g(\x)}{\partial x_i} & \text{for } i=j \\
       -x_j \frac{\partial g(\x)}{\partial x_i} & \text{for } i \neq j
    \end{cases}
    \label{eq:proof:derivate}
\end{equation}
In LayerNorm~\cite{ba2016layer}, we assume for simplicity $\mathbb{E}[\x] = 0$, then the partial derivative simplifies to
\begin{align*}
    & \mathbb{V}[\x] = \mathbb{E}[\x^2] - \mathbb{E}[\x]^2 = \mathbb{E}[\x^2] \\
    & \frac{\partial \mathbb{V}[\x]}{\partial x_i} = \frac{2}{N} x_i
\end{align*}
Further, the partial derivative of RMSNorm~\cite{zhang2019root} is
\begin{equation*}
    \frac{\partial \text{RMSNorm}}{\partial x_i} = \frac{x_i}{\sqrt{N \sum_k x_k}}
\end{equation*}

At reference point $\tilde{x}_i = 0$, the diagonal elements in Equation~\eqref{eq:proof:derivate} $i \neq j$ are zero, yielding the Taylor decomposition:
\begin{equation*}
    f_i(\x) = \frac{\partial f_i}{\partial x_i}\Bigr|_{\tilde{x}_i=0} x_i  + \tilde{b}_i = \frac{x_i}{\varepsilon} + \tilde{b}_i 
\end{equation*}
To enforce a strict notion of the conservation property~\eqref{lrp:conservation}, the bias term $\tilde{b}_i$  can be excluded or evenly distributed across the input variables. Because we have only a single input variable, the bias can be considered as part of $x_i$.
\begin{equation}
    R_i^{l-1} = \left( \frac{x_i}{\varepsilon} + \tilde{b}_i \right) \frac{R_i^l}{\frac{x_i}{\varepsilon} + \tilde{b}_i}
\end{equation}
Since there is only one input variable and one output, the decomposition is equivalent to the identity function, as discussed in the Section~\ref{methods:non_linear} about component-wise non-linearities. Thus, we conclude that the identity rule applies in this case.
\begin{equation}
    R_i^{l-1} = R_i^{l}
\end{equation}
Note, that this rule is numerically stable because $f_j(0)=0$ as discussed in Section~\ref{app:bias}.

\subsubsection{Violation of the Conservation Property in bi-linear Matrix Multiplication}
\label{app:proof:conservation}

In the following we proof that the application of the \mbox{$\varepsilon$-rule~\eqref{eq:epsilon} without the uniform rule~\eqref{eq:uniform}} on bi-linear matrix multiplication violates the conservation property~\eqref{lrp:conservation}. We reiterate and generalize the Lemma 3 of \cite{chefer2021transformer} which establishes that $0$-LRP ($\varepsilon=0$) violates conservation.

Recall, that matrix multiplication is defined as:
\begin{equation*}
     \textbf{O}_{jp} = \sum_i  \textbf{A}_{ji} \textbf{V}_{ip}
\end{equation*} 
The $\varepsilon$-rule is the solution to applying a linearization~\eqref{eq:taylor} to a linear layer~\eqref{eq:linear}. For computing relevance values for $\textbf{A}_{ji}$ using the $\varepsilon$-rule, we treat $\textbf{V}_{ip}$ as a constant weight matrix with zero bias, and similarly for attributing $\textbf{A}_{ji}$.
The derived relevance propagation rules are given by:
\begin{align*}
    & \tilde{R}_{ji}^{l-1}(\textbf{A}_{ji}) = \sum_p \textbf{A}_{ji} \textbf{V}_{ip} \frac{R_{jp}^l}{\textbf{O}_{jp} + \varepsilon} \\
    & \tilde{R}_{ip}^{l-1}(\textbf{V}_{ip}) = \sum_j \textbf{A}_{ji} \textbf{V}_{ip} \frac{R_{jp}^l}{\textbf{O}_{jp} + \varepsilon} \\
\end{align*}
The conservation property~\eqref{lrp:conservation} states, that the total relevance at layer $l$ must be equal to the total relevance at layer $l-1$.

The total relevance at layer $l$ is given by
\begin{equation*}
      R^l = \sum_{j,p} R_{jp}^l
\end{equation*}
and the total relevance at layer $l-1$ is computed by:
\begin{equation*}
     R^{l-1} = \sum_{j,i} \tilde{R}_{ji}^{l-1} + \sum_{i,p} \tilde{R}_{ip}^{l-1} = 2 \sum_{j,p} \frac{\textbf{O}_{jp}}{\textbf{O}_{jp} + \varepsilon} R_{jp}^l \approx 2 R^l
\end{equation*}
with $\varepsilon \ll |\textbf{O}_{jp}|$. This results in a violation of the conservation property as $R^l \neq R^{l-1}$. However, by employing Proposition 3.3~\eqref{prop:matrix_multiplication}, that is a sequential application of the $\varepsilon$-rule and uniform rule~\eqref{eq:uniform}, we ensure conservation by dividing with the factor $2$:
\begin{equation*}
     R^{l-1} = \sum_{j,i} R_{ji}^{l-1} + \sum_{i,p} R_{ip}^{l-1} = \sum_{j,p} \frac{\textbf{O}_{jp}}{\textbf{O}_{jp} + \varepsilon} R_{jp}^l \approx R^l
\end{equation*}
It is evident that $\varepsilon$ absorbs a negligible proportion of the relevance to safeguard numerical stability. The proof is also valid for the $z^+$-rule~\eqref{eq:appendix:zplus-rule}, where only positive contributions are taken into consideration.

\newpage
\section{Appendix II: Experimental Details}
In the following sections, we provide additional details about the experiments performed.

\subsection{Models and Datasets}
For ImageNet faithfulness, we utilized the pretrained Vision Transformer B-16, L-16 and L-32  weights of the PyTorch model zoo~\cite{paszke2019pytorch}. We randomly selected 3200 samples (fixed set for all baselines) such that the standard error of mean converges to below 1\% of the mean value.

For Wikipedia and IMDB faithfulness, we evaluated the pretrained LLaMa 2-7b hosted on huggingface~\cite{wolf2019huggingface} on 4000 randomly selected validation dataset samples (fixed set for all baselines). For SQuAD\,v2, we utilize the pretrained Flan-T5 and Mixtral 8x7b weights hosted on huggingface.
Further, for Wikipedia next word prediction we evaluated the model on a context size of 512 (from beginning of article until context length is reached), while the context size in SQuAD\,v2 varies between 169 to 4060. Although Flan-T5 was trained on a smaller context size of 2000 tokens, the relative positional encoding allows it to handle longer context sizes with at least 8192 tokens~\cite{shaham2023zeroscrolls}. 

All computations are performed in the Brain Floating Point (bfloat16) half-precision format to save memory consumption. bfloat16 trades precision for a higher dynamic range than standard float16, and hence prevents numerical errors due to overflow. In this regard, the impact of quantized number formats on (Attn)LRP attributions remains a topic to be investigated. In addition, all linear weights in Mixtral 8x7b are quantized to the 4 bit integer format using \textit{bitsandbytes}~\cite{dettmers2024qlora} (but computation still performed in bfloat16).

SQuAD\,v2 encompasses numerous questions that are either unanswerable or subject to incorrect predictions by the model. Consequently, only instances where the model accurately predicts the correct response are considered. Additionally, only the testset is utilized to mitigate a potential overfitting bias during the training phase, if applicable.
Finally, for SQuAD\,v2 top-1 accuracy and IoU, we utilized the following prompt:

\texttt{Context: [text of dataset sample] Question: [question of dataset sample] Answer:}

Flan-T5 does not require a system prompt, while for Mixtral 8x7b we use before the context the system prompt:

\texttt{Use the context to answer the question. Use few words.}

Because Flan-T5 typically provides the correct answer directly, we explain the first token of the answer only. Conversely, Mixtral 8x7b generates full sentences; within these, we identify the positions of the answer tokens and explain all tokens that constitute the correct answer only. To achieve this, we calculate heatmaps for each answer token and add these heatmaps to produce the final heatmap. For gradient-based methods, this process can be parallelized by initiating the backward pass at the designated token positions with the logit output for LRP and with the value 1 for all other baselines, while initializing the remaining output tokens with zero.

For IMDB, we added a last linear layer to a frozen LLaMa 2-7b model and finetuned only the last layer, which achieves 93\% accuracy on the validation dataset. 

If we encountered NaN values for a sample, we removed it from the evaluation. This happened for Grad $\times$ AttnRollout and AtMan in the Wikipedia dataset. However, the standard error of the mean remains small, as can be seen in Table~\ref{tab:faithfulness}.

\subsection{Input Perturbation Metrics}\label{sec:input_perturbation}
In the following, we summarize the perturbation process introduced by \cite{samek2016evaluating} in a condensed manner. 

Given an attributions map $R_i^l(x_i)$ per input features $\x = \{x_i\}_{i=1}^N$ in layer $l$. $\mathcal{H}$ denotes a set of relevance values for all input features $x_i$:
\begin{equation}
    \mathcal{H} = (R_{0}^0(x_{0}), R_{1}^0(x_{1}), ..., R_{N-1}^0(x_{N-1}))
\end{equation} 
Then, the \emph{flipping} perturbation process iteratively substitutes input features with a baseline value $\textbf{b} \in \mathbb{R}^{N}$ (the baseline might be zero, noise generated from a Gaussian distribution, or pixels of a black image in the vision task).
Another reverse equivalent variant, referred as \emph{insertion}, begins with a baseline $\textbf{b}$ and reconstructs the input \x step-wise. The function performing the perturbation is denoted by $\textbf{g}^{F}$ for flipping and $\textbf{g}^{I}$ for insertion. The perturbation procedure is either conducted in a MoRF (Most Relevant First) or LeRF (Least Relevant First) manner based on the sorted members of $\mathcal{H}$.
Regardless of the replacement function, the MoRF and LeRF perturbation processes can be defined as recursive formulas at step $k = \{0, 1, ..., N-1\}$:

\begin{equation*}
\text{MoRF Pert. Process}=
\begin{cases}
\x_{MoRF}^{0} = \x\\
\x_{MoRF}^{k} = \textbf{g}^{(F|I)}(\x_{MoRF}^{k-1}, \textbf{b})\\
\x_{MoRF}^{N-1} = \textbf{b}
\end{cases}
\end{equation*}
where $\x_{MoRF}^{k}$ denotes the perturbed input feature $\x$ at step $k$ in MoRF process.

\begin{equation*}
\text{LeRF Pert. Process}=
\begin{cases}
\x_{LeRF}^{0} = \textbf{b}\\
\x_{LeRF}^{k} = \textbf{g}^{(F|I)}(\x_{LeRF}^{k-1}, \textbf{b})\\
\x_{LeRF}^{N-1} = \x
\end{cases}
\end{equation*}
where $\x_{LeRF}^{k}$ denotes the perturbed input feature $\x$ at step $k$ in LeRF process.

Results of these processes are perturbed input sets of $\mathcal{X}^{F}_{MoRF}=(\x_{MoRF}^{0}, \x_{MoRF}^{1}, ...,\x_{MoRF}^{N-1})$ and $\mathcal{X}^{F}_{LeRF}=(\x_{LeRF}^{0}, \x_{LeRF}^{1}, ...,\x_{LeRF}^{N-1})$. By feeding these sets to the model and computing the corresponding logit output $f_j$, a curve will be induced and consequently the area $A$ under the curve can be calculated:

\begin{align}
    A^{F}_{MoRF} = A^{I}_{LeRF} &= \frac{1}{N} \sum^{N-1}_{k=0} f_j(\x_{MoRF}^{k})
\end{align}
where $\x_{MoRF}^{k} \in \mathcal{X}^{F}_{MoRF}$ or $\x_{MoRF}^{k} \in \mathcal{X}^{I}_{LeRF}$.

It is notable that the area below the least relevant order insertion curves are identical to the most relevant order flipping curves and that the area below the least relevant order flipping curves are identical to the most relevant order insertion curves. Hence, by using $\mathcal{X}^{I}_{MoRF}$, $A^{I}_{MoRF}=A^{F}_{LeRF}$ can be computed similarly. A faithful explainer results in a low value of $A^{F}_{MoRF}$ or $A^{I}_{LeRF}$.  
Further, a faithful explainer is expected to have large $A^{F}_{LeRF}$ or $A^{I}_{MoRF}$ values. 

Ultimately to reduce introducing out-of-distribution manipulations and the sensitivity towards the chosen baseline value, the work of \cite{blucher2024decoupling} proposes to leverage both insights and to obtain a robust measure as 
\begin{align*}
    &\Delta A^{F} = A^{F}_{LeRF} - A^{F}_{MoRF} \\
   & \Delta A^{I} = A^{I}_{MoRF} - A^{I}_{LeRF}
\end{align*}
where a higher score signifies a more faithful explainer.

We performed all faithfulness perturbations with a baseline value of zero. In the case of LLMs, we aggreated the relevance for each token and flipped the entire embedding vector of input tokens to the baseline value. For ViTs, we used the relevances of the input pixels and flipped input pixels to the baseline value.

\begin{figure*}
    \centering
        \includegraphics[width=0.99\linewidth]{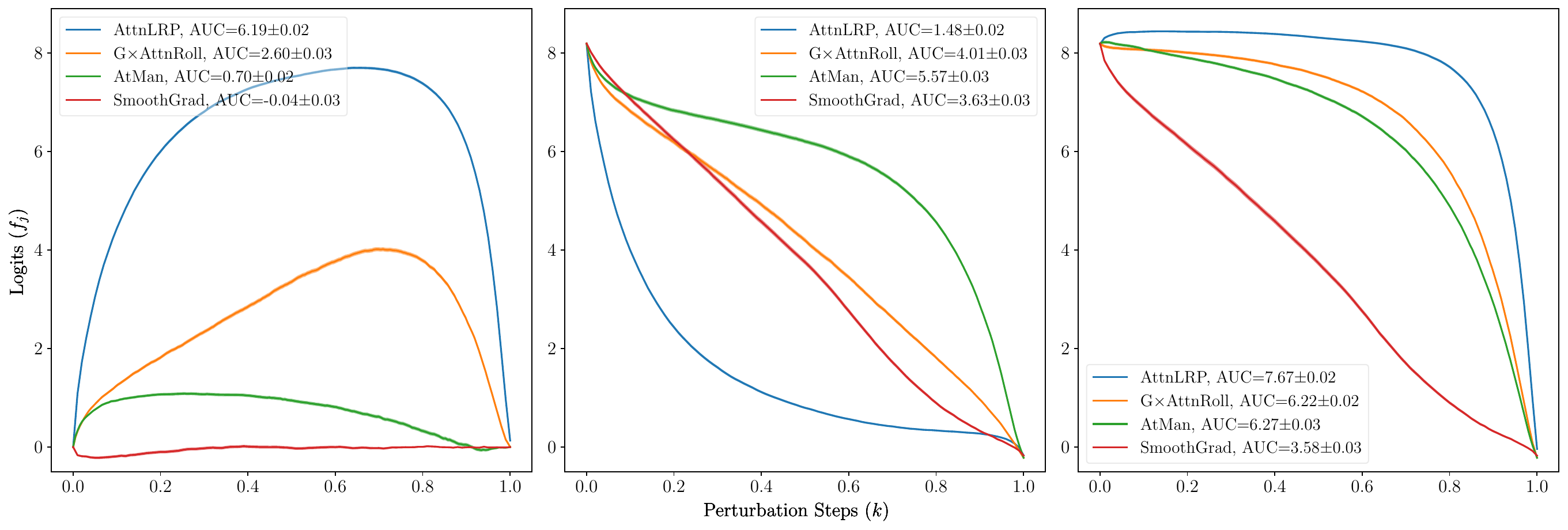}

    \caption{Comparison of the \gls{ours} (ours) with the $\gamma$-rule, Grad$\times$AttnRoll \cite{chefer2021generic}, AtMan \cite{deb2023atman}, and SmoothGrad \cite{smilkov2017smoothgrad} techniques through the perturbation experiment (faithfulness) on the ViT-B-16 using 3200 random samples of ImageNet. From left to right, the plots correspond to $f_{j}(\mathcal{X}^{F}_{LeRF})-f_{j}(\mathcal{X}^{F}_{MoRF})$ (large area is good), $f_{j}(\mathcal{X}^{F}_{MoRF})$ (steep decline is good), and $f_{j}(\mathcal{X}^{F}_{LeRF})$ (slow decline is good). ``AUC" denotes the Area under Curve.
    }
    \label{app:fig:auc_comparison}
\end{figure*}

\subsection{Hyperparameter search for Baselines}\label{app:sweepsearch_sg_atman}

As describes in Appendix~\ref{app:details:baselines}, several baseline attribution methods have hyperparameters that must be tuned to the datasets. The default parameters are described in Appendix~\ref{app:details:baselines}, and to reduce the search space, we optimize a subset of the hyperparameters.
The hyperparameters of SmoothGrad ($\sigma \in [0.01, 0.25]$), AtMan ($\text{suppression value}\in[0.1, 1.0]$ and threshold $\in[0, 1.0]$), AttnRoll ($\text{discard threshold}\in[0.90, 1.00]$), and G$\times$AttnRoll ($\text{discard threshold}\in[0.90, 1.00]$) are selected to be optimized. The used hyperparameters for the perturbation experiments are available in the captions of Tables \ref{tab:app:vit_more}, \ref{app:table_vit_imagenet}, \ref{app:tab:llm_wiki} and \ref{app:tab:llm_imdb}.

For LLMs, we have not noticed a significant impact on the heatmaps for different discard threshold values of G$\times$AttnRoll. For AttnRoll, the impact is minimal. Hence, we choose the default value of 1 (nothing is discarded, as proposed in the original works~\cite{abnar2020quantifying, chefer2021generic}).

Regarding SQuAD\,v2, we set AtMan's $p=0.7$ for Mixtral 8x7b and $p=0.9$ for Flan-T5-XL.
For SmoothGrad, we set $\sigma=0.1$ for Mixtral 8x7b and Flan-T5-XL.

\subsection{Impact of Model Architectural Choices on AttnLRP Performance}
\label{app:ablations}
We evaluated \gls{ours} on three model classes that incorporate different types of layers. 

\underline{Flan-T5}: This encoder-decoder architecture employs self-attention and cross-attention layers \eqref{eq:ap:attention}. The FFN layers are a sequential application of linear layers with GELU non-linearities inbetween.

\underline{LLaMa 2}: This decoder architecture utilizes only self-attention layers \eqref{eq:ap:attention}. However, in the FFN layers, we have an additional element-wise non-linear weighting with a SiLU non-linearity \eqref{eq:ap:ffn}.

\underline{Mixtral 8x7b}: This mixture of experts model uses self-attention layers \eqref{eq:ap:attention} and FFN layers with non-linear weighting \eqref{eq:ap:ffn} like the LLaMa 2. In addition, there are FFN routing layers with a softmax weighting \eqref{eq:ap:routing}.

\begin{align}
    & \text{Attention: } \text{Softmax} (\textbf{W}_q \ \x \ (\textbf{W}_k \ \x)^\top) \ \textbf{W}_v \ \x \label{eq:ap:attention}\\
    & \text{FFN $\times$ Non-Linearity: } \text{SiLU}(\textbf{W}_1 \ \x) \odot \textbf{W}_2 \ \x \label{eq:ap:ffn}\\
    & \text{Routing: } \sum_i \text{Softmax(TopK(} \textbf{W}_g \ \x))_i \ \text{FFN}_i(\x) \label{eq:ap:routing}
\end{align}
where $\textbf{W}_q$, $\textbf{W}_k$, $\textbf{W}_v$, $\textbf{W}_1$, $\textbf{W}_2$, $\textbf{W}_g$ are linear weight parameters, ($\odot$) is element-wise multiplication, $i \in \mathbb{N}$ the number of expert FFN layers and TopK is returning the top-k elements.

In Table~\ref{tab:app:ablation}, we study the impact of AttnLRP rules \wrt CP-LRP on all three different layer types \eqref{eq:ap:attention}, \eqref{eq:ap:ffn} and \eqref{eq:ap:routing}. We start as baseline with all rules of CP-LRP on all layer types, then we successively substitute CP-LRP rules with AttnLRP rules for specific layer types.

For CP-LRP, we use the rules described in Appendix~\ref{app:difference}. In the original work of \cite{ali2022xai}, FFN layers weighted with non-linearities and routing layers are not discussed. Analogously to the argumentation in their work, we regard the non-linearity as constant weight and attribute only through the FFN path using the $\varepsilon$-rule.

As demonstrated in Table~\ref{tab:app:ablation}, the application of AttnLRP rules enhances the performance across all layers, regardless of their type. Moreover, the rate of improvement increases with the number of non-linearities present in the model.

\begin{table*}[h!]
    \centering
    \caption{This Table is an extension of Table~\ref{tab:faithfulness}: Faithfulness scores as area between the least and most relevant order perturbation curves on LLaMa 2 alongside the top-1 accuracy and IoU in parenthesis for Flan-T5 and Mixtral 8x7b. We start as baseline with all rules of CP-LRP on all layer types, then we successively substitute CP-LRP rules with AttnLRP rules for specific layer types if they exist in the model. We observe that AttnLRP's improvement is not confined to the attention mechnism alone, but all layers that contain operations that are not attributable with other LRP variants. The more complex the architecture, the better the performance of AttnLRP compared to CP-LRP.}
    \label{tab:app:ablation}    
    \begin{tabular}{@{}l@{\hspace{0.3em}}c@{\hspace{0.3em}}c@{\hspace{0.3em}}c@{\hspace{0.3em}}c@{\hspace{0.7em}}c@{}}
    
        \toprule
        \textbf{Method on Layer}  & \multicolumn{2}{c}{\textbf{LLaMa 2-7b}} & \textbf{Mixtral 8x7b} & \textbf{Flan-T5-XL}\\
         &  IMDB $\uparrow$ & Wikipedia $\uparrow$ & SQuAD\,v2 $\uparrow$ & SQuAD\,v2 $\uparrow$\\
        \midrule

        \textbf{Baseline (All Layers)} &  &  &  &  \\

        CP-LRP & $1.72$ & $7.85$ & $0.50 \ (0.40)$ & $0.90 \ (0.83)$ \\

        \midrule

        \textbf{+ Attention Mechanism} &  &  &  &  \\

        \gls{ours} & $2.09$ & $9.49$ & $0.70 \ (0.53)$ & $0.94 \ (0.84)$ \\

        \midrule

        \textbf{+ FFN $\times$ Non-Linearity} &  &  &  &  \\

        \gls{ours} & $2.50$ & $10.93$ & $0.78 \ (0.57)$ & - \\

        \midrule

        \textbf{+ Routing Layer} &  &  &  &  \\

        \gls{ours} & - & - & $0.96 \ (0.72)$ & - \\
        
        \bottomrule
    \end{tabular}
\end{table*}

\subsection{LRP Composites for ViT}\label{app:lrp_composite}

Applying the $\varepsilon$-rule on all linear layers inside LLMs is sufficient to obtain faithful and noise-free attributions. However, for the vision transformers, we apply the $\gamma$-rule on all linear layers (including the convolutional layers) outside the attention module. Since the $\gamma$-rule has a hyperparameter, the work~\cite{pahde2023optimizing} proposed to tune the parameter using a grid-search. 
This optimization search (or in an LRP context known as composite search) is computational highly demanding.

The vision transformer consists of many linear layers. Our proposed approach is to use different $\gamma$ values across different layer types.

\newpage

According to \cite{vaswani2017attention} the attention module consists of several linear layers which we refer to as \textit{LinearInputProjection}.
\begin{equation*}
    \textbf{Q} = \textbf{W}_q \textbf{X} + \textbf{b}_q
\end{equation*}
\begin{equation*}
    \textbf{K} = \textbf{W}_k \textbf{X} + \textbf{b}_k
\end{equation*}
\begin{equation*}
    \textbf{V} = \textbf{W}_v \textbf{X} + \textbf{b}_v
\end{equation*}
In the attention layer, after the softmax~(\ref{eq:attention}), there exists another linear layer performing the output projection back into the residual stream, denoted as \textit{LinearOutputProjection}:
\begin{equation*}
    \textbf{y} = \textbf{W}_o \textbf{O} + \textbf{b}_o
\end{equation*}

The other layers in the whole network, will be referred to as \textit{Linear}.

The perturbation experiment had been conducted over these layers using different types of rules including Epsilon, ZPlus, Gamma, and AlphaBeta (with $\alpha=2$ and $\beta=1$ according to \cite{montavon2019layer}).

\begin{figure}[t]
  \centering
  \includegraphics[width=1\linewidth]{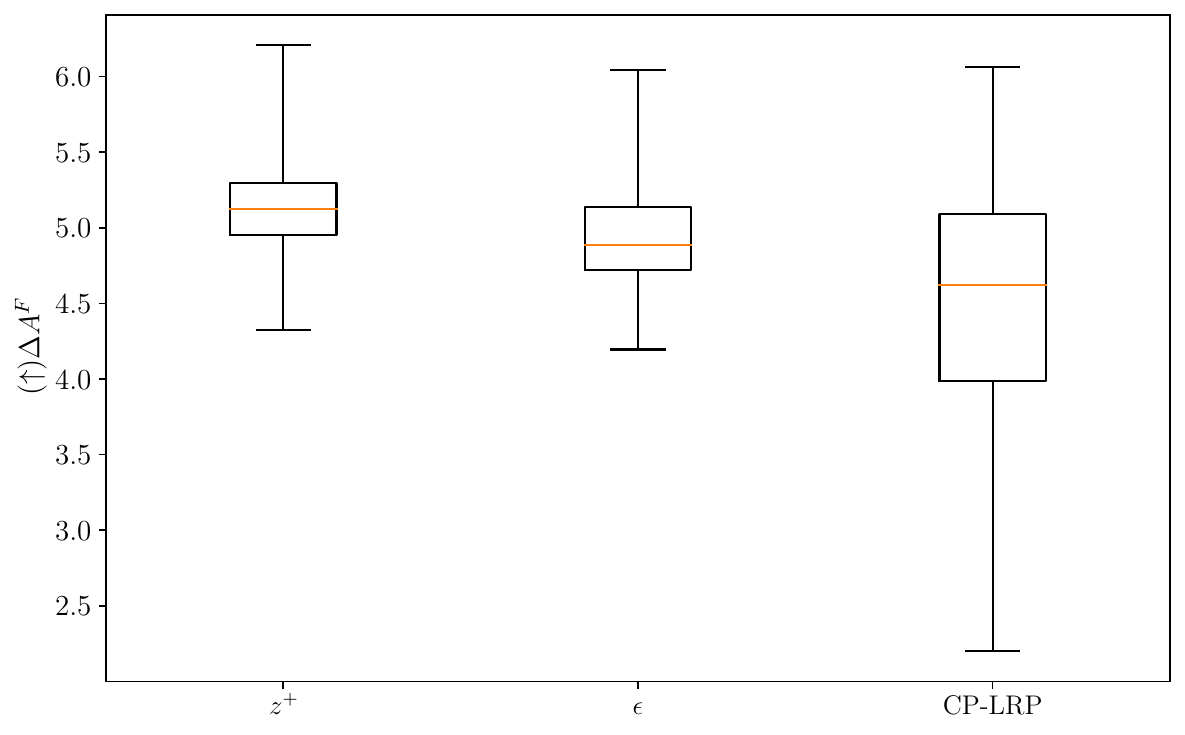}
   \caption{Statistics on Rules used for \textit{softmax} layers: Either applying $z^+$, $\varepsilon$-rule, or regarding as constant as proposed in CP-LRP.
   Propagating relevance values through (specifically by applying $z^+$ rule) \textit{softmax} improves the faithfulness of explanations compared to the case where we block its propagation.}
\label{app:fig:vit_pf_softmax}
\end{figure}

\begin{figure}[t]
  \centering
  \includegraphics[width=1\linewidth]{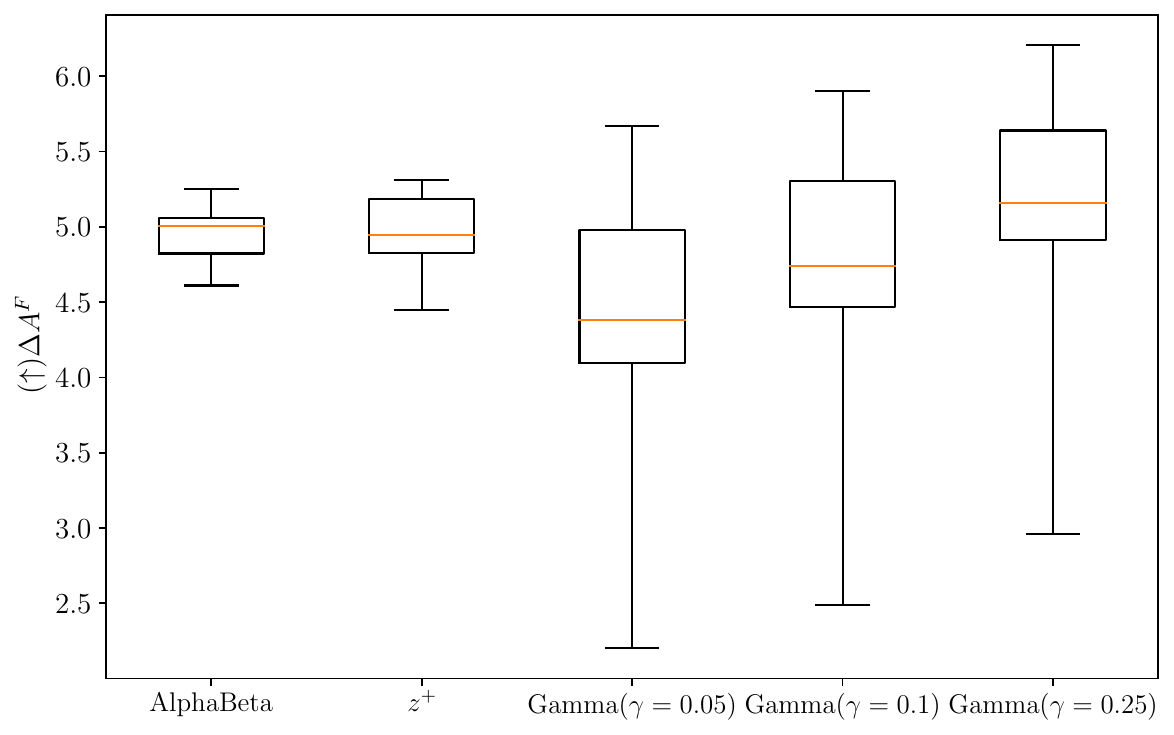}
   \caption{Statistics on Rules used for \textit{Convolution} layers: Applying $z^+$ and AlphaBeta proposes acceptable results however the most faithful results can be reached via Gamma($\gamma=0.25$).}
\label{app:fig:vit_pf_conv}
\end{figure}

\begin{figure}[t]
  \centering
  \includegraphics[width=1\linewidth]{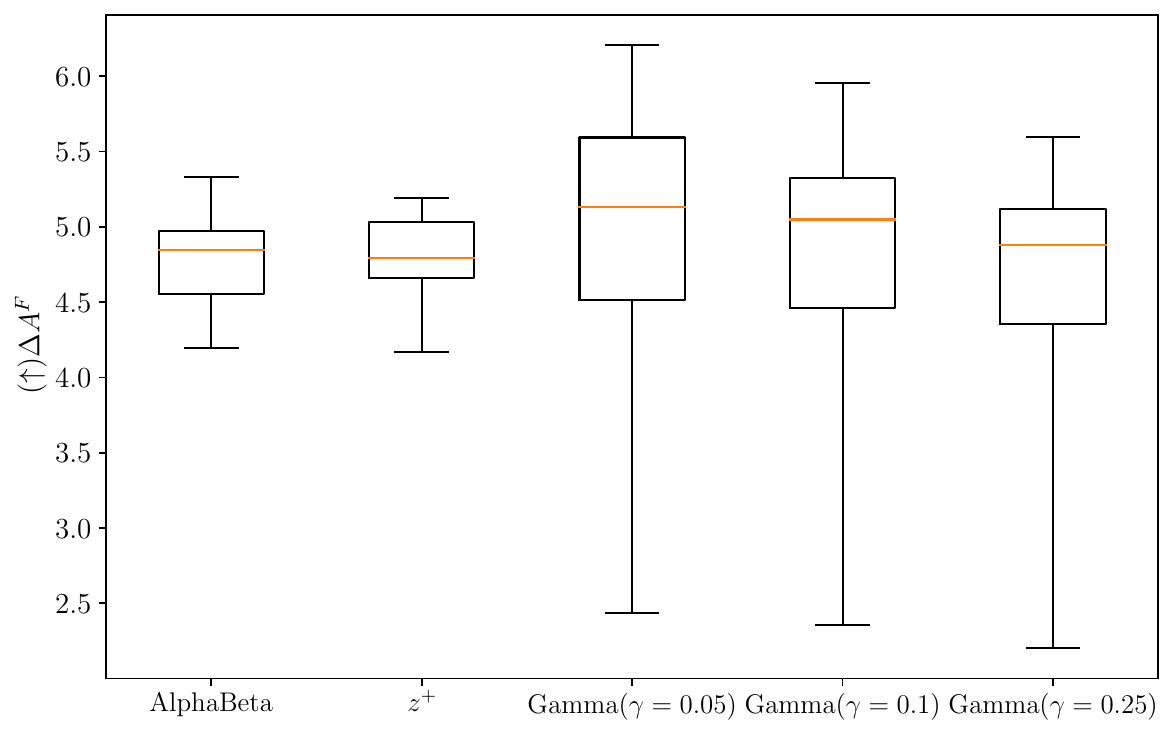}
   \caption{Statistics on Rules used for \textit{Linear} layers: Similar to \textit{Convolution} layers, Gamma seems more promising however with different $\gamma$ value ($0.05$ in this case).}
\label{app:fig:vit_pf_linear}
\end{figure}

\begin{figure}[t]
  \centering
  \includegraphics[width=1\linewidth]{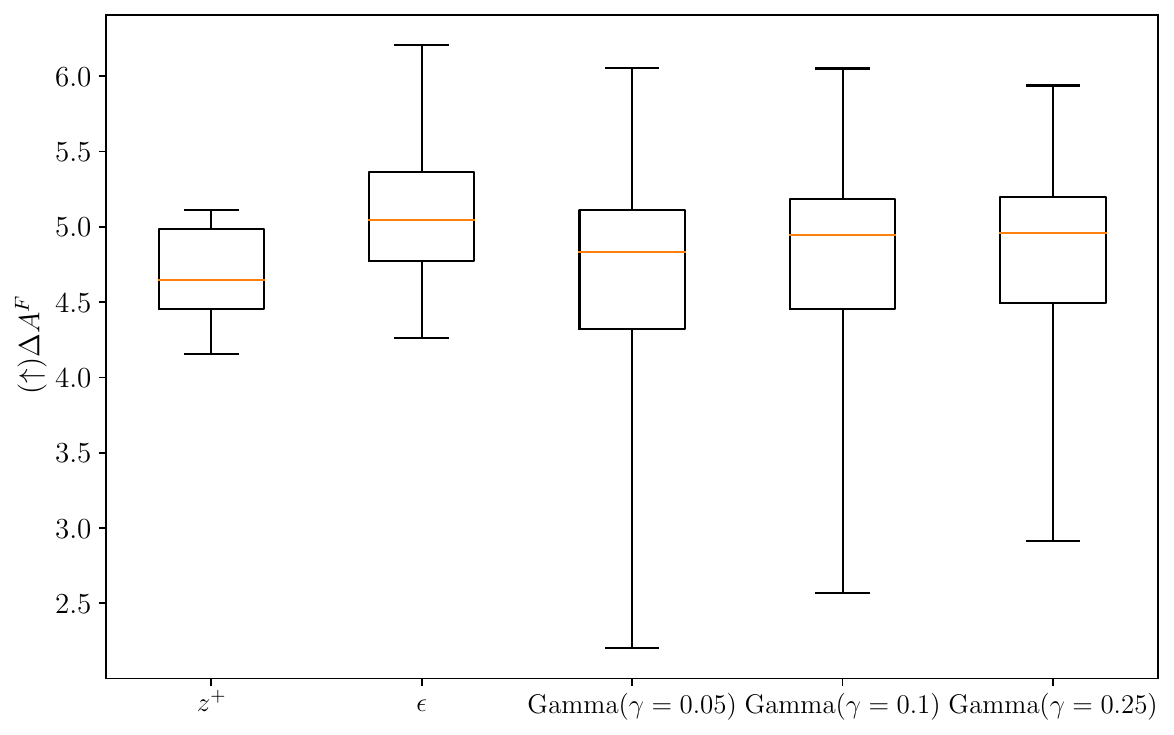}
   \caption{Statistics on Rules used for \textit{LinearInputProjection} layers: Gamma and $\epsilon$ rules are competitive in this case, however since there is larger difference between the minimum and the lower quartile in Gamma rules, the most faithful choice will be $\epsilon$-rule.}
   \label{app:fig:vit_pf_linearinputprojection}
\end{figure}

\begin{figure}[t]
  \centering
  \includegraphics[width=1\linewidth]{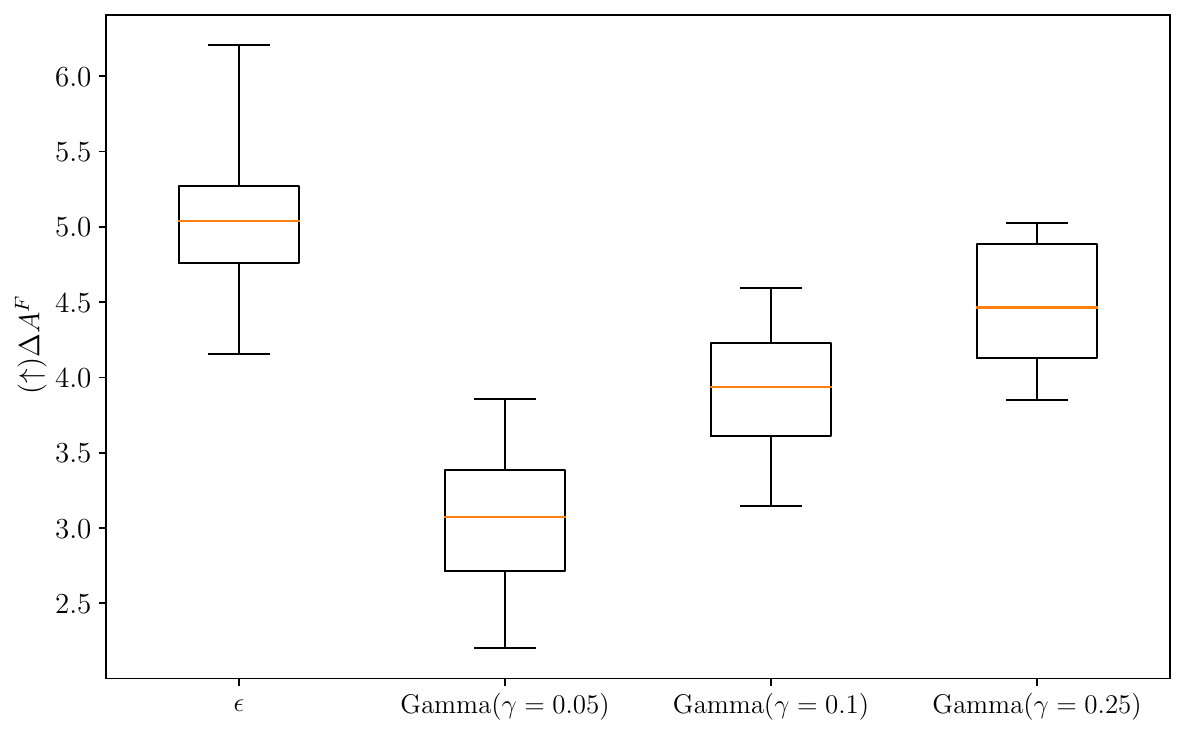}
   \caption{Statistics on Rules used for \textit{LinearOutputProjection} layers: The $\epsilon$-rule outperforms other rules  clearly.}
\label{app:fig:vit_pf_linearoutprojection}
\end{figure}

The most faithful composite, that we obtained for \gls{ours} and CP-LRP,
is in Table \ref{app:tab:vit_composite}.
More details over the statistics of the conducted experiments are available in Figures \ref{app:fig:vit_pf_softmax}, \ref{app:fig:vit_pf_conv}, \ref{app:fig:vit_pf_linear}, \ref{app:fig:vit_pf_linearinputprojection}, \ref{app:fig:vit_pf_linearoutprojection}.

\begin{table}[h!]
    \centering
    \caption{Proposed composite for the \gls{ours} and CP-LRP methods used for the Vision Transformer.}
    \label{app:tab:vit_composite}
    \begin{tabular}{lc}
        \toprule
        Layer Type & Rule Proposed \\
        \midrule
         Convolution & Gamma($\gamma=0.25$) \\
         Linear & Gamma($\gamma=0.05$) \\
         LinearInputProjection & Epsilon \\
         LinearOutputProjection & Epsilon \\
        \bottomrule
    \end{tabular}
\end{table}

\begin{figure*}
    \centering
        \includegraphics[width=1\linewidth]{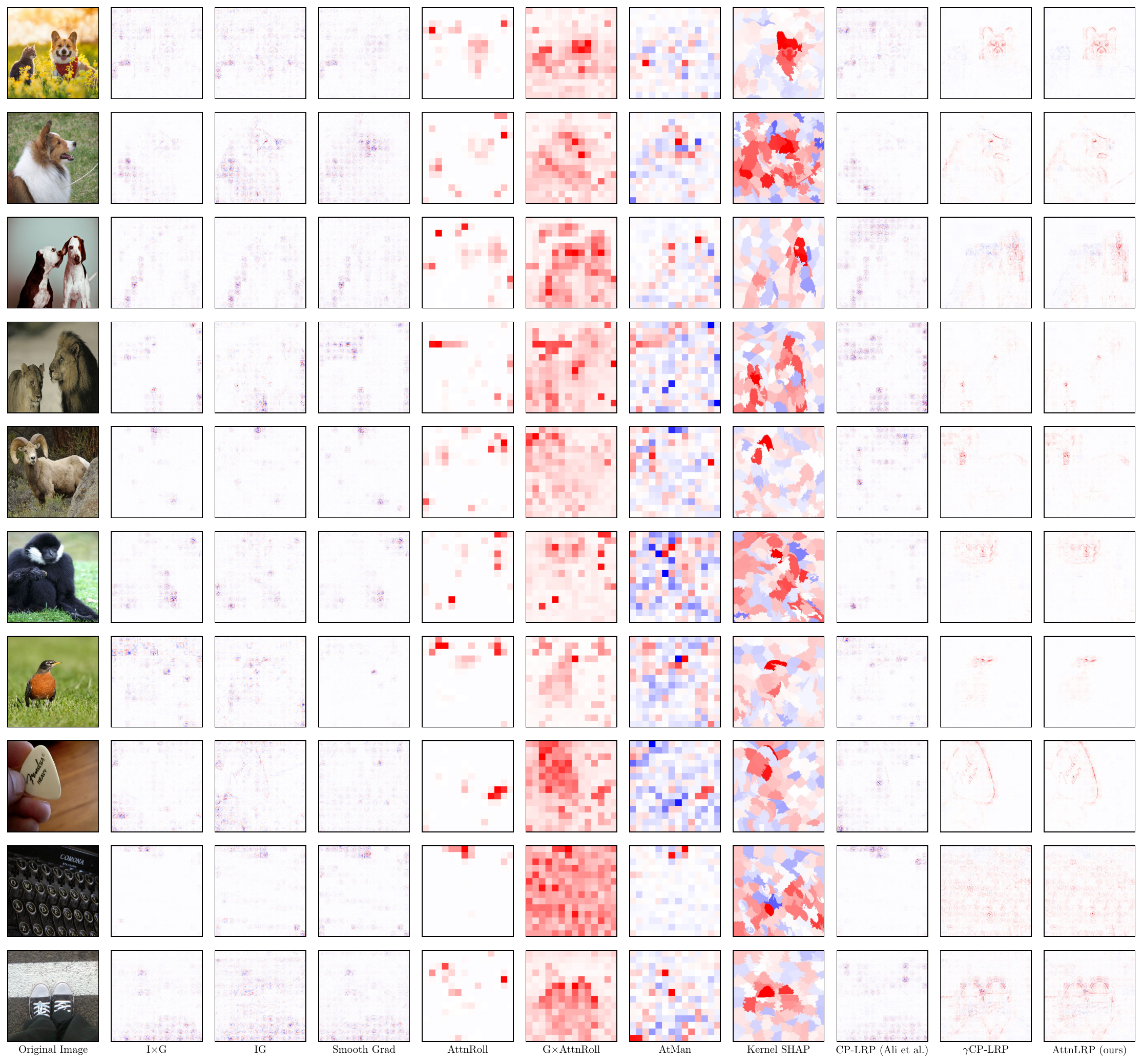}

    \caption{Explanation heatmaps of the methods used for the perturbation experiments on the vision transformer ViT-B-16. A checkerboard effect is visible for almost every method, especially in AttnRoll \cite{abnar2020quantifying}, G$\times$AttnRoll \cite{chefer2021generic}, and AtMan \cite{deb2023atman}. We improve upon CP-LRP \cite{ali2022xai} by applying the $\gamma$-rule as described in \ref{app:lrp_composite}.
    While qualitatively $\gamma$CP-LRP (with $\gamma$ extension for ViT) and AttnLRP give similar explanations, quantitative results in Table \ref{tab:faithfulness} and Table \ref{tab:app:vit_more} show a consistent improvement of AttnLRP over $\gamma$CP-LRP in terms of faithfulness. Moreover, attributing query and key linear layers within the attention layer is possible with AttnLRP only, while it is not possible with CP-LRP. We leave these further explorations for future work.}
    \label{fig:vit_heatmaps}
\end{figure*}

\subsection{Additional Perturbation Evaluations on Vision Transformers}
\label{app:extra:vit}
Table~\ref{tab:app:vit_more} presents additional perturbation results for the vision transformers ViT-L-16 and ViT-L-32 evaluated on ImageNet. Our method surpasses all comparative baselines, while the proposed enhancement, $\gamma$CP-LRP (which applies the $\gamma$-rule across all linear layers for CP-LRP~\cite{ali2022xai}), remains highly competitive. In more complex model architectures that incorporate a greater variety of non-linearities, our method demonstrates more superiority, as elaborated in Appendix~\ref{app:ablations}.
Tuning the $\gamma$-parameter for $\gamma$CP-LRP and \gls{ours} in a grid-search (see Appendix~\ref{app:lrp_composite}) resulted for both models in the same composite described in Table~\ref{app:tab:vit_composite}. However, there is no assurance that other models share the same $\gamma$ parameters.

\begin{table*}[h!]
    \centering
    \caption{Faithfulness scores as area between the least and most relevant order perturbation curves \cite{blucher2024decoupling} for ViT-L-16 and ViT-L-32 on ImageNet. For ViT-L-16, we set for SmoothGrad $\sigma=0.01$, for AtMan $p=1.0$ and $t=0.1$, for AttnRoll $dt=0.90$, and for G$\times$AttnRoll $dt=0.92$. For ViT-L-32, we set for SmoothGrad $\sigma=0.01$, for AtMan $p=1.0$ and $t=0.1$, for AttnRoll $dt=0.95$, and for G$\times$AttnRoll $dt=1.0$.
    }
    \label{tab:app:vit_more}    
    \begin{tabular}{@{}l@{\hspace{0.3em}}c@{\hspace{0.3em}}c@{\hspace{0.3em}}c@{\hspace{0.3em}}c@{\hspace{0.7em}}c@{}}
    
        \toprule
        \textbf{Method} & \textbf{ViT-L-16} & \textbf{ViT-L-32} \\
         & ImageNet $\uparrow$ & ImageNet $\uparrow$ \\
        \midrule

        Random & \phantom{-}\phantom{-}$0.01{\,\pm\, 0.01}$ & \phantom{-}\phantom{-}$0.01{\,\pm\, 0.01}$ \\
        \midrule
        
        Input$\times$Grad~\cite{simonyan2014deep} & \phantom{-}\phantom{-}$1.20\tiny{\,\pm\, 0.06}$ & \phantom{-}\phantom{-}$0.98\tiny{\,\pm\, 0.07}$ \\

        IG~\cite{sundararajan2017axiomatic} & \phantom{-}\phantom{-}$0.96\tiny{\,\pm\, 0.07}$ & \phantom{-}\phantom{-}$1.45\tiny{\,\pm\, 0.06}$\\

        SmoothGrad~\cite{smilkov2017smoothgrad} & $-0.10\tiny{\,\pm\, 0.01}$ & $-0.09\tiny{\,\pm\, 0.04}$  \\
        \midrule
        
        GradCAM~\cite{chefer2021transformer} & \phantom{-}\phantom{-}$0.19\tiny{\,\pm\, 0.06}$ & \phantom{-}\phantom{-}$2.21\tiny{\,\pm\, 0.08}$  \\%

        AttnRoll~\cite{abnar2020quantifying} & \phantom{-}\phantom{-}$1.41\tiny{\,\pm\, 0.08}$ & \phantom{-}\phantom{-}$1.90\tiny{\,\pm\, 0.07}$ \\

        Grad$\times$AttnRoll~\cite{chefer2021generic} & \phantom{-}\phantom{-}$2.86\tiny{\,\pm\, 0.06}$ & \phantom{-}\phantom{-}$2.69\tiny{\,\pm\, 0.06}$ \\

        \midrule
        AtMan~\cite{deb2023atman} & \phantom{-}\phantom{-}$1.58\tiny{\,\pm\, 0.08}$ & \phantom{-}\phantom{-}$0.09\tiny{\,\pm\, 0.05}$ \\ 

        KernelSHAP~\cite{lundberg2017Shap} & \phantom{-}\phantom{-}$4.35\tiny{\,\pm\, 0.04}$ & \phantom{-}\phantom{-}$4.90\tiny{\,\pm\, 0.03}$ \\
        \midrule

        CP-LRP ($\varepsilon$-rule, \citet{ali2022xai}) & \phantom{-}\phantom{-}$4.96\tiny{\,\pm\, 0.05}$ & \phantom{-}\phantom{-}$4.07\tiny{\,\pm\, 0.05}$ \\

        CP-LRP ($\gamma$-rule for ViT, as proposed here) & \phantom{-}\phantom{-}$6.97\tiny{\,\pm\, 0.04}$ &  \phantom{-}\phantom{-}$5.99\tiny{\,\pm\, 0.04}$ \\

        \gls{ours} (ours) &  \phantom{-}\phantom{-}$\textbf{7.17}\tiny{\,\pm\, 0.04}$  & \phantom{-}\phantom{-}$\textbf{6.06}\tiny{\,\pm\, 0.04}$ \\

        \bottomrule
    \end{tabular}
\end{table*}

\subsection{Attributions on SQuAD\,v2}
\label{app:compare_squad}
\begin{figure*}[h!] 
  \centering
  \includegraphics[width=1\linewidth]{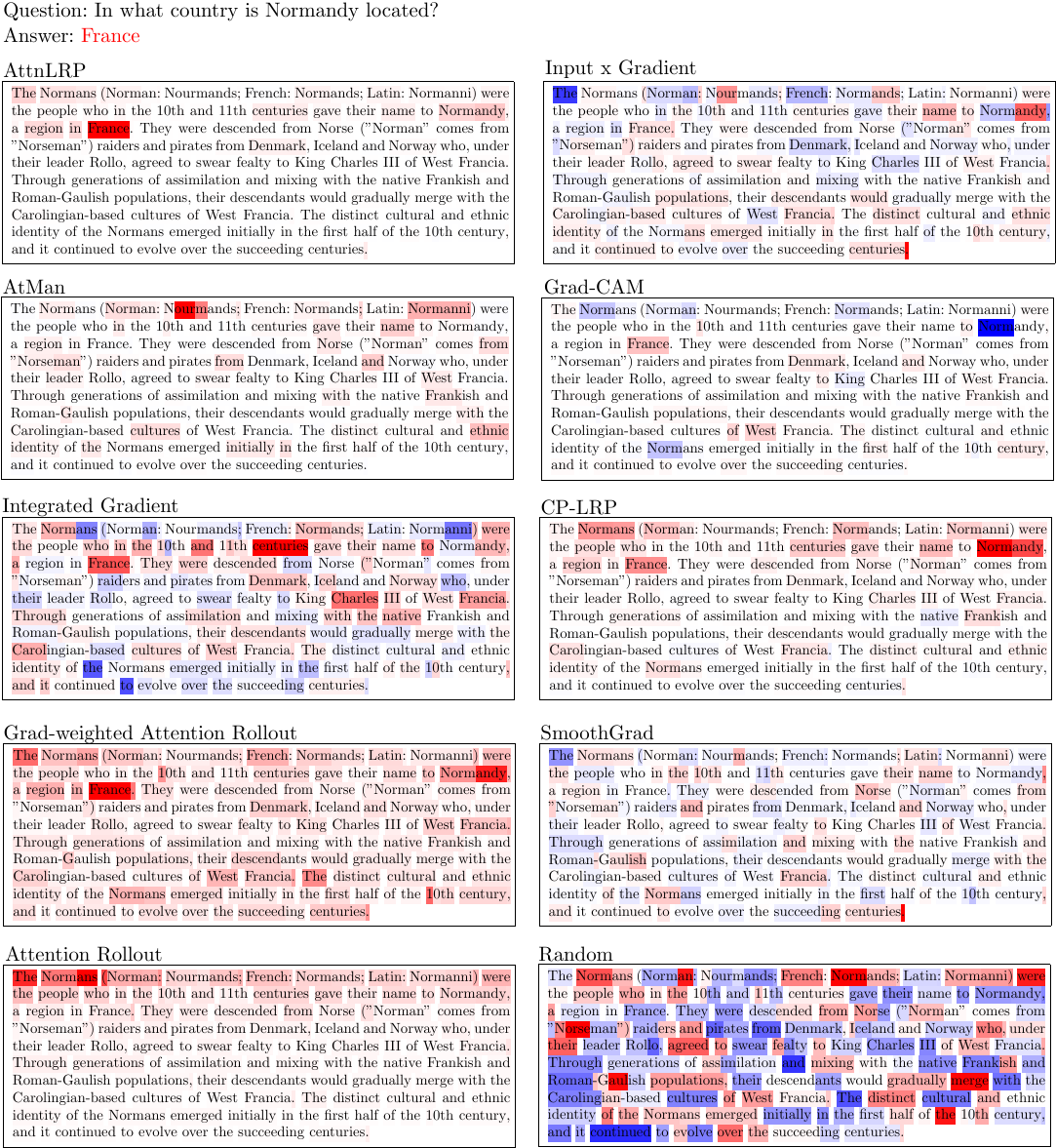}
   \caption{Evaluation on the Mixtral 8x7b model: We compute attributions for different state-of-the-art methods for the answer token ``France". Gradient-based methods such as G$\times$I, SmoothGrad, IG or Grad-CAM are noisy. Grad$\times$Attn Rollout suffers from background noise. While AtMan usually generates sparse heatmap, in this case it fails (compare Figure~\ref{app:fig:compare_squad}). CP-LRP highlights ``Normandy" the strongest, while AttnLRP highlights the correct token ``France''. For comparison, we also visualize a random attribution with Gaussian noise.}
\label{app:fig:france}
\end{figure*}

\begin{figure*}[h!] 
  \centering
  \includegraphics[width=0.8\textwidth]{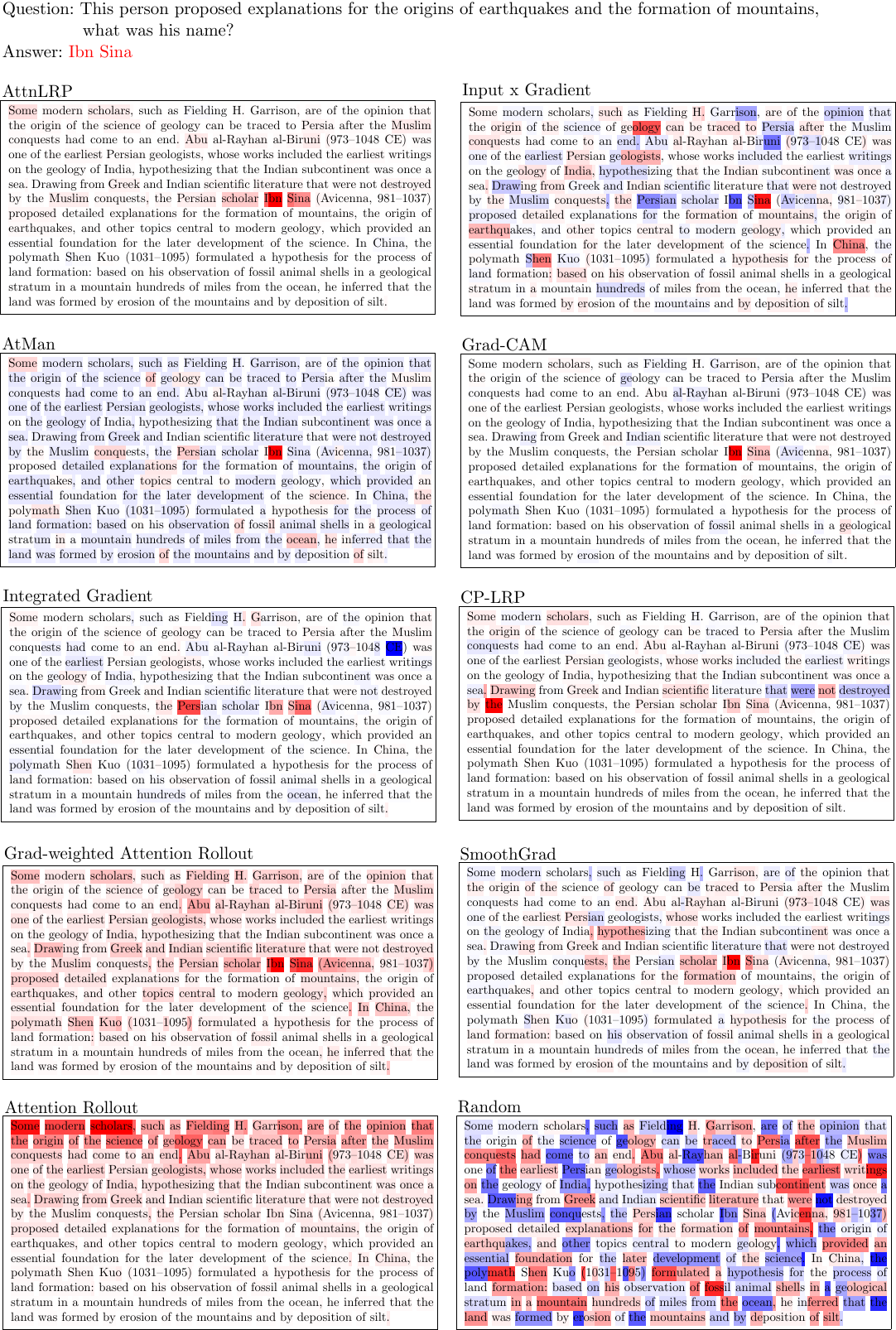}
   \caption{Evaluation on the Mixtral 8x7b model: We compute attributions for different state-of-the-art methods for all tokens at the same time inside the answer ``Ibn Sina". Gradient-based methods such as G$\times$I, SmoothGrad and IG are noisy. Grad-CAM highlights the correct tokens except it misses the beginning token ``I" of the word ``Ibn". Likewise AtMan fails to highlight all tokens. Grad$\times$Attn Rollout suffers from background noise. CP-LRP resembles random noise, while AttnLRP highlights the correct tokens ``Ibn Sina'' in its entirety. For comparison, we also visualize a random attribution with Gaussian noise.}
\label{app:fig:geology}
\end{figure*}

\begin{figure*}[h!] 
  \centering
  \includegraphics[width=1\linewidth]{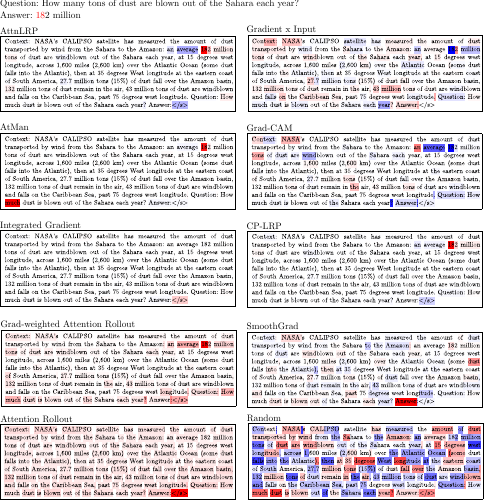}
   \caption{Evaluation on the Flan-T5-XL model: We compute attributions for different state-of-the-art methods on the first token of the answer (highlighted in red). Gradient-based methods such as G$\times$I, SmoothGrad, IG or Grad-CAM are noisy. Grad$\times$Attn Rollout suffers from background noise. AtMan produces highly sparse attributions, assigning an equal amount of relevance to a token, which is part of the question, as to token \texttt{18}. CP-LRP has a different weighting of the tokens \eg the word `much' in the question is not highlighted by CP-LRP, while AttnLRP highlights it stronger and AtMan focuses excessively on it. For comparison, we also visualize a random attribution with Gaussian noise.}
\label{app:fig:compare_squad}
\end{figure*}

In Figure~\ref{app:fig:france} and Figure~\ref{app:fig:geology}, we illustrate attributions on the Mixtral 8x7b for different state-of-the-art methods on the SQuAD\,v2 dataset. In Figure~\ref{app:fig:compare_squad}, we depict attributions for Flan-T5-XL. For comparison, we also visualize a random attribution with Gaussian noise.

The similarity between \gls{ours} and CP-LRP in Flan-T5-XL are in line with the quantitative evaluation from Table \ref{tab:faithfulness}, which shows a small, but consistent advantage of \gls{ours} over CP-LRP wrt.\ top-1 accuracy, while in Mixtral 8x7b and LLaMa 2, AttnLRP substantially outperforms, which is also visible in the heatmaps. This is due to the different number of non-linearities present in the models: Flan-T5-XL consists only of standard attention layers, while LLaMa 2 and Mixtral 8x7b have additional FFN layers with non-linear weighting or routing layers making the attribution process more difficult for CP-LRP. This effect is studied in Appendix~\ref{app:ablations}.
In general, gradient-based methods such as G$\times$I, SmoothGrad, IG and Grad-CAM are noisy and often not informative. Attention Rollout  and Grad$\times$Attention Rollout suffer from background noise. While the performance of AtMan is in some cases excellent as in Figure~\ref{app:fig:compare_squad}, the method fails in other cases as in Figure~\ref{app:fig:france}.

In Figure~\ref{app:fig:france} and \ref{app:fig:geology} for Mixtral 8x7b, most methods fail to highlight the correct answer tokens most strongly, except \gls{ours}, confirming the quantitative evaluation from Table \ref{tab:faithfulness}.

In Figure~\ref{app:fig:compare_squad}, the heatmaps of I$\times$G or Grad-CAM seem to be inverted, hence we experimented with inverting the attributions on a subset, however we did not notice improvement and applied the rules with their original definition. AtMan produces highly sparse attributions, assigning large positive relevance to the answer token \texttt{18}, however, also assigning a similar amount of relevance to the token \texttt{much}, which is part of the question. \gls{ours} and CP-LRP identify the token \texttt{18} as being the most relevant token and also relate it (by assigning positive and negative relevance) to other information in the text such as \texttt{27.7},  \texttt{132} or \texttt{average}. We conjecture that such targeted contrasting reflects the reasoning process of the model (\eg, is necessary to distinguish between related questions about how many tons are blown out vs.\ how many tons remain on the ground). A systematic analysis of these effects remain an interesting topic for future work.

\subsection{Benchmarking Cost, Time and Memory Consumption}
\label{app:benchmarking}
We benchmark the runtime and peak GPU memory consumption for computing a single attribution for LLaMa 2 with batch size 1 on a node with four A100-SXM4 40GB, 512 GB CPU RAM and 32 AMD EPYC 73F3 3.5 GHz. Because AtMan, LRP and AttnRollout-variants need access to the attention weights, we did not use flash-attention~\cite{dao2022flashattention}.

\begin{figure*}
    \centering
        \includegraphics[width=0.99\linewidth]{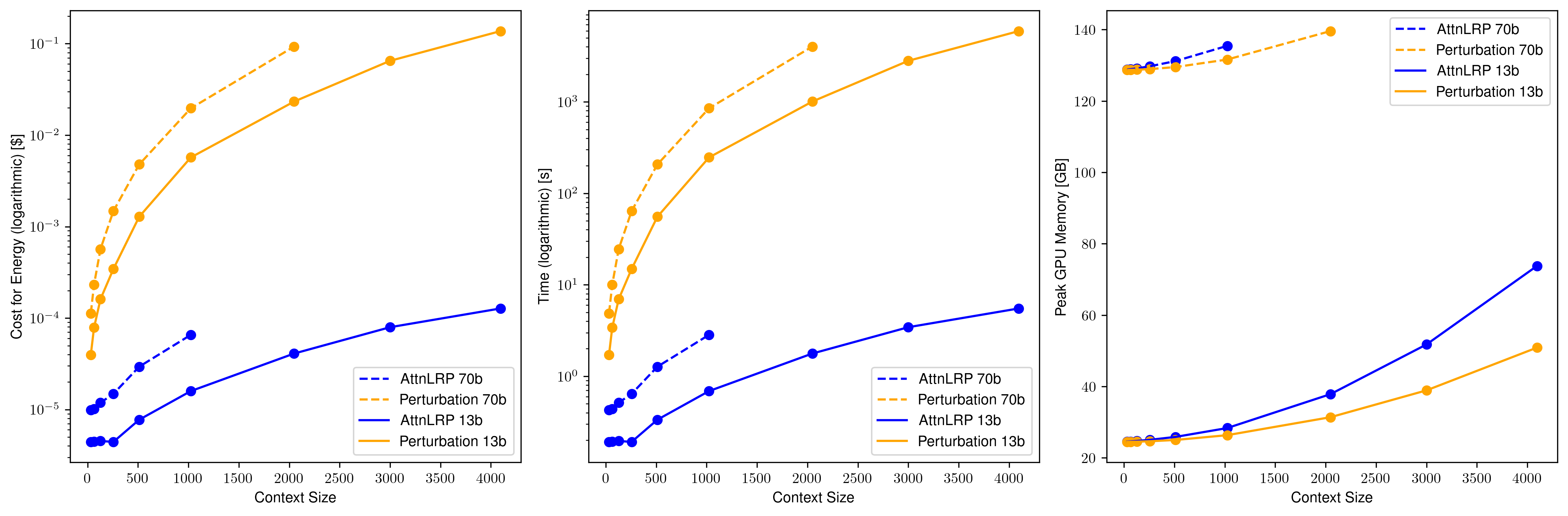}

    \caption{From left to right: Cost in dollar, time in seconds and peak GPU memory in gigabytes for AttnLRP and linear-time perturbation. Evaluated on LLaMa 2-70b and LLaMa 2-13b models on a node with four A100-SXM4 40GB. G$\times$AttnRollout is in the range of AttnLRP and omitted for clarity of visualization. Because AttnLRP consumes more than 160 GB of RAM, the curves for the 70b model stop. Measured at fixed intervals of context size 32, 64, 128, 256, 512, 1024, 2048, 3000, 4096.
    }
    \label{app:fig:costs}
\end{figure*}

To calculate energy cost, we assume a price of $0.16$\,\$ per kWh of energy, and that a single A100 GPU consumes on average 130W. Figure~\ref{app:fig:costs} depicts the cost, the runtime and peak GPU memory consumption. Since perturbation-based methods are memory efficient, a 70b model with full context size of 4096 is attributable. However, LRP with checkpointing requires more memory than a node supplies. 

\subsection{Attributions of Knowledge Neurons}

\begin{figure}[hbt!]
  \centering
  \includegraphics[width=1\linewidth]{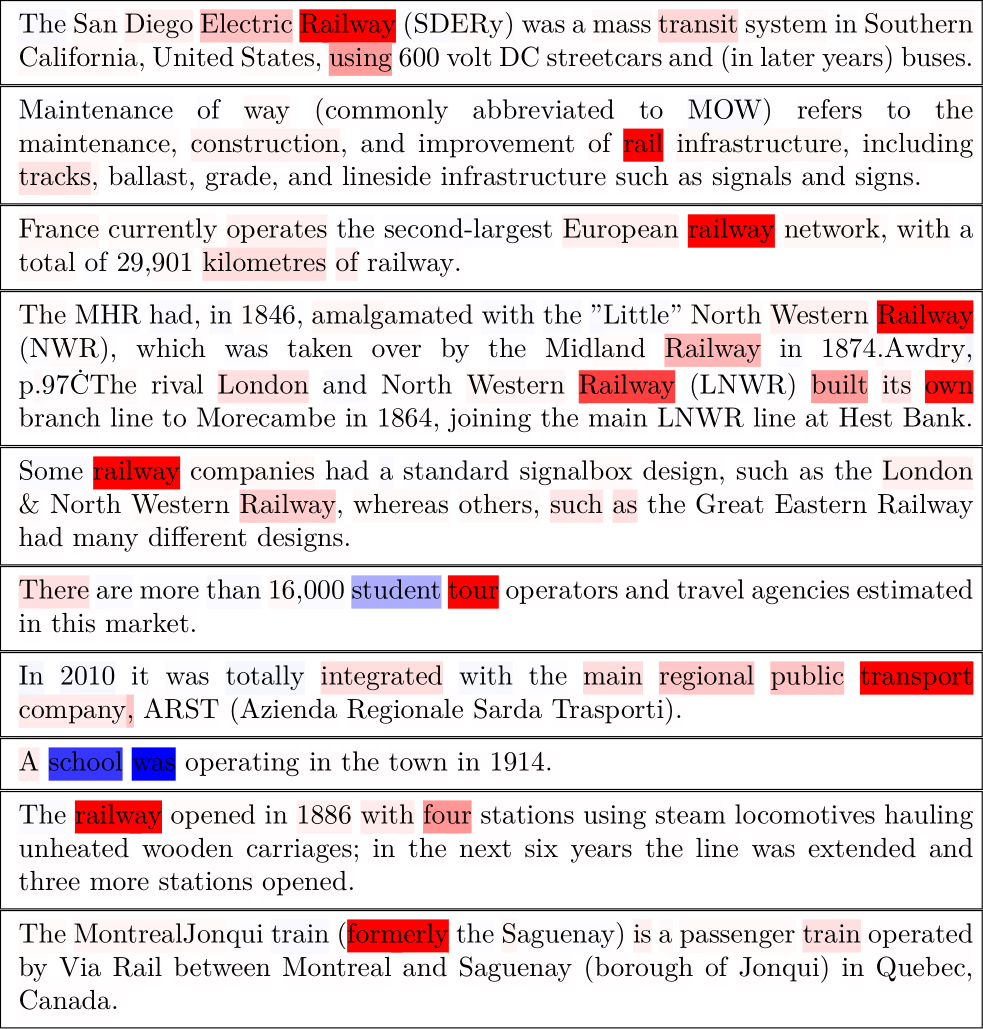}
   \caption{AttnLRP attributions on top 10 \gls{amax} sentences collected over the Wikipedia summary dataset for neuron \texttt{\#256}, in \texttt{layer 18}. The knowledge neuron seems to activate for transport systems (railways in particular).}
\label{app:fig:knowledgeneuron1}
\end{figure}

\begin{figure}[hbt!]
  \centering
  \includegraphics[width=1\linewidth]{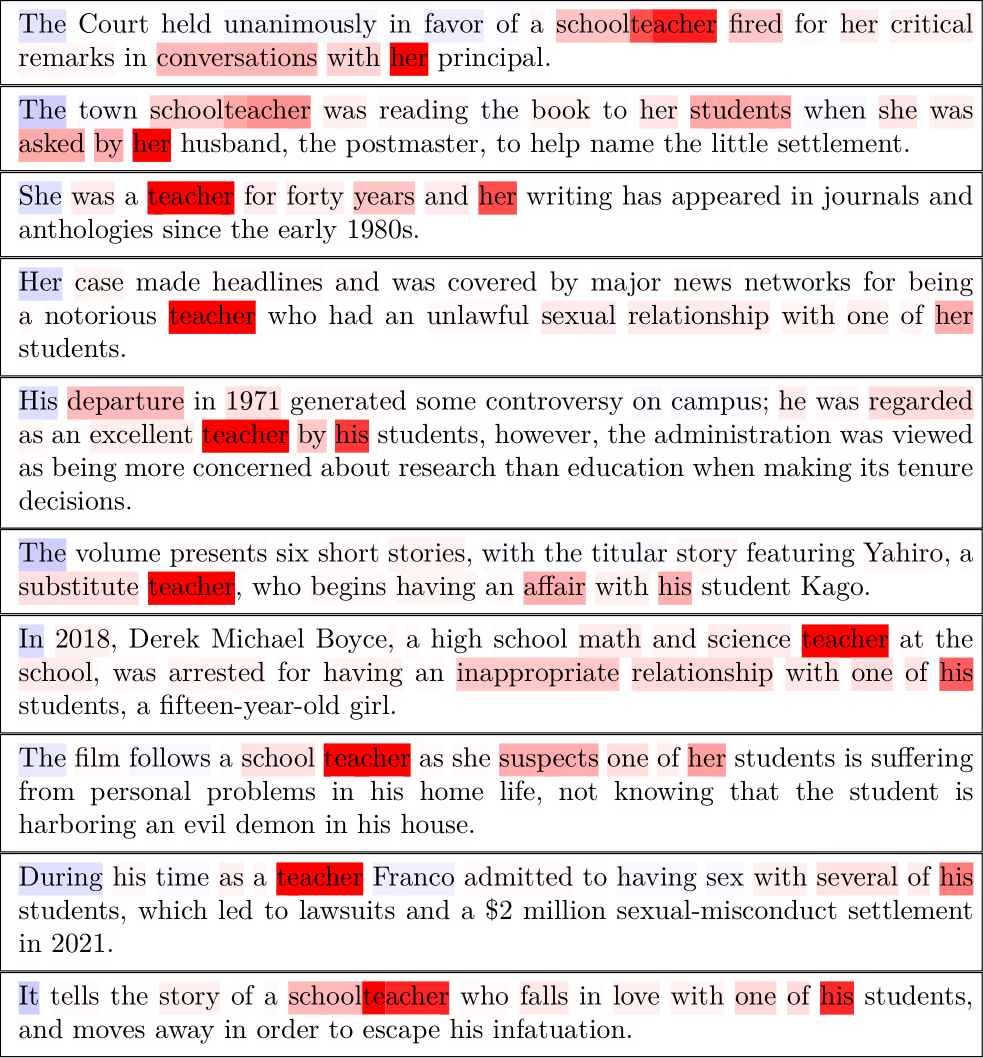}
   \caption{AttnLRP attributions on top 10 \gls{amax} sentences collected over the Wikipedia summary dataset for neuron \texttt{\#2207}, in \texttt{layer 20}. The knowledge neuron is activating for `teacher', in unusual context such as inappropriate behavior, sexual misconduct etc.}
\label{app:fig:knowledgeneuron2}
\end{figure}

\begin{figure}[hbt!]
  \centering
  \includegraphics[width=1\linewidth]{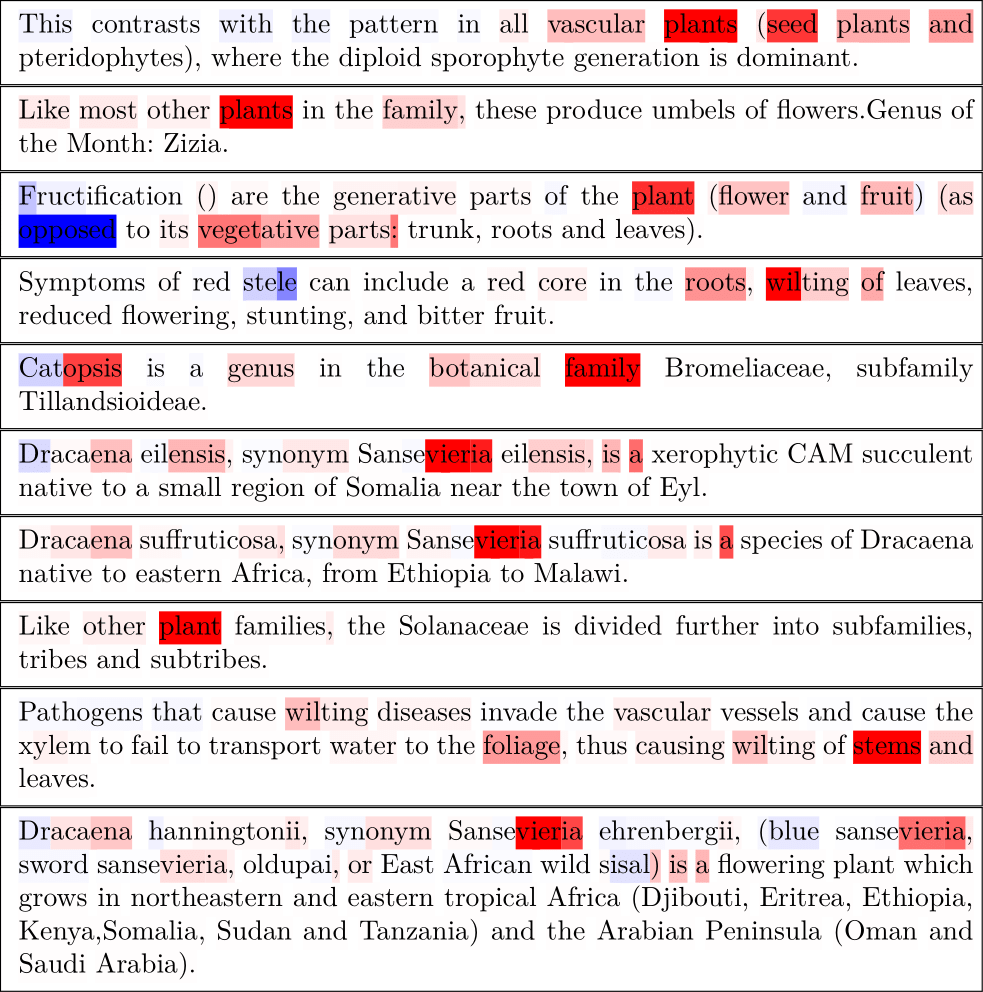}
   \caption{AttnLRP attributions on top 10 \gls{amax} sentences collected over the Wikipedia summary dataset for neuron \texttt{\#922}, in \texttt{layer 18}. The knowledge neuron seems to be activating for scientific descriptions of plants.}
\label{app:fig:knowledgeneuron3}
\end{figure}

Figure~\ref{app:fig:knowledgeneuron1}, \ref{app:fig:knowledgeneuron2} and \ref{app:fig:knowledgeneuron3} illustrate the top 10 sentences in the Wikipedia summary dataset that maximally activate a knowledge neuron. We applied \gls{ours} to highlight the tokens inside these reference samples. We observe that knowledge neurons exhibit remarkable disentanglement, \eg, neuron \texttt{\#256} of \texttt{layer 18} shown in Figure~\ref{app:fig:knowledgeneuron1} seems to encode concepts related to transport systems (railways in particular), while neuron \texttt{\#2207} of \texttt{layer 20} shown in Figure~\ref{app:fig:knowledgeneuron2} seems to encode the concept teacher, in particular a teacher, in an unusual context (\eg, inappropriate behavior, sexual misconduct). The degree of disentanglement should be studied in future work.

\begin{table*}[h!]
    \centering
    \caption{ViT-B-16 Perturbation Experiment (Faithfulness). For SmoothGrad, we set $\sigma=0.01$, for AtMan $p=1.0$ and $t=0.1$, for AttnRoll $dt=0.99$, and for G$\times$AttnRoll $dt=0.91$.
    ``all epsilon" indicates that the $\varepsilon$-rule has been used on the linear and convolutional layers. The term ``best" refers to the utilization of LRP with the composite proposed in \ref{app:lrp_composite}. $\Delta A^{F}$ denotes the area under the curve for a \emph{flipping} perturbation experiment which leverages both $A_{MoRF}^{F}$ of the most relevant first order, and $A_{LeRF}^{F}$ of least relevant first order. ($\Delta A^{F}=A_{LeRF}^{F}-A_{MoRF}^{F}$). As discussed in Section~\ref{sec:input_perturbation}, this is equivalent to \emph{insertion} perturbation.}
    \label{app:table_vit_imagenet}\label{app:tab:vit_faithfulness}
    \begin{tabular}{lccc}
        \toprule
        \textbf{Methods}  & \multicolumn{3}{c}{\textbf{ViT-B-16}}\\
         & \multicolumn{3}{c}{ImageNet}\\
         & ($\uparrow$)$\Delta A^{F}$ & ($\downarrow$)$A_{MoRF}^{F}$ & ($\uparrow$)$A_{LeRF}^{F}$\\
        \midrule

        Random & 0.01 & 4.71 & 4.71 \\ 

        I$\times$G & 0.90 & 2.78 & 3.69 \\ 

        IG & 1.54 & 2.55 & 4.10 \\

        SmoothG & -0.04 & 3.63 & 3.58 \\ 

        GradCAM & 0.27 & 5.35 & 5.63 \\ 

        AttnRoll & 1.31 & 4.866 & 6.17 \\ 

        G$\times$AttnRoll & 2.60 & 4.01 & 6.22\\ 

        AtMan & 0.70 & 5.57 & 6.27\\ 

        CP-LRP (all epsilon) & 2.53 & 2.45 & 4.98\\

        $\gamma$CP-LRP (best) & 6.06 & 1.53 & 7.59\\

        \gls{ours} (all epsilon) & 2.79 & 5.22 & 2.42\\

        \gls{ours} (best) & \textbf{6.19} & \textbf{1.48} & \textbf{7.67} \\ 

        \bottomrule
    \end{tabular}
\end{table*}

\begin{table*}[h!]
    \centering
    \caption{Wikipedia Perturbation Experiment (Faithfulness). For SmoothGrad, we set $\sigma=0.1$, for AtMan $p=1.0$, for AttnRoll $dt=1$, and for G$\times$AttnRoll $dt=1$.
    "all epsilon" indicates that the $\varepsilon$-rule has been used on all linear layers. $\Delta A^{I}$ denotes the area under the curve for the \emph{insertion} perturbation experiment which leverages both $A_{MoRF}^{I}$ of the most relevant first order, and $A_{LeRF}^{I}$ of least relevant first order. ($\Delta A^{I}=A_{MoRF}^{I}-A_{LeRF}^{I}$). As discussed in Section~\ref{sec:input_perturbation}, this is equivalent to \emph{flipping} perturbation.
    }
    \label{app:tab:llm_wiki}
    \begin{tabular}{lccc}
        \toprule
        \textbf{Methods}  & \multicolumn{3}{c}{\textbf{LLaMa 2-7b}}\\
         & \multicolumn{3}{c}{Wikipedia}\\
         
         & ($\uparrow$)$\Delta A^{I}$ & ($\uparrow$)$A_{MoRF}^{I}$ & ($\downarrow$)$A_{LeRF}^{I}$\\
        \midrule

        Random & -0.07 & 2.31 & 2.38 \\ 

        I$\times$G & 0.18 & 1.27 & 1.09 \\ 

        IG & 4.05 & 3.74 & -0.31 \\ 

        SmoothG & -2.22 & 0.68 & 2.90\\  

        GradCAM & 2.01 & 2.36 & 0.35\\ 

        AttnRoll & -3.49 & 1.46 & 4.95\\ 

        G$\times$AttnRoll & 9.79 & 8.79 & -1.00\\ 

        AtMan & 3.31 & 4.06 & 0.76\\ 

        CP-LRP (all epsilon) & 7.85 & 6.43 & -1.42\\ 

        \gls{ours} (all epsilon) & \textbf{10.93} & \textbf{9.08} & \textbf{-1.85}\\ 

        \bottomrule
    \end{tabular}
\end{table*}

\begin{table*}[h!]
    \centering
    \caption{IMDB Perturbation Experiment (Faithfulness), For SmoothGrad we set  $\sigma=0.05$, for AtMan $p=0.7$, for AttnRoll $dt=1$, and for G$\times$AttnRoll $dt=1$.
     "all epsilon" indicates that the $\varepsilon$-rule has been used to propagate relevance to the layers. $\Delta A^{I}$ demonstrates the area under the curve for the perturbation experiment of the type \textit{Insertion} which leverages insights from both $A_{MoRF}^{I}$ of the most relevant first order, and $A_{LeRF}^{I}$ of least relevant first order.  ($\Delta A^{I}=A_{MoRF}^{I}-A_{LeRF}^{I}$). As discussed in Section~\ref{sec:input_perturbation}, this is equivalent to \emph{flipping} perturbation.}
     \label{app:tab:llm_imdb}
    \begin{tabular}{lccc}
        \toprule
        \textbf{Methods}  & \multicolumn{3}{c}{\textbf{LLaMa 2-7b}}\\
         & \multicolumn{3}{c}{IMDB}\\
         & ($\uparrow$)$\Delta A^{I}$ & ($\uparrow$)$A_{MoRF}^{I}$ & ($\downarrow$)$A_{LeRF}^{I}$\\
        \midrule

        Random & -0.01 & -0.47 & -0.46 \\ 

        I$\times$G & 0.12 & -0.69 & -0.81 \\ 

        IG & 1.23 & -0.06 & -1.29 \\  

        SmoothG & 0.25 & -0.74 & -0.98\\ 

        GradCAM & -0.82 & -1.10 & -0.28\\ 

        AttnRoll & -0.64 & -0.64 & 0.00\\  

        G$\times$AttnRoll & 1.61 & 0.77 & -0.84\\  

        AtMan & -0.05 & -0.54 & -0.49 \\ 

        CP-LRP (all epsilon) & 1.72 & 0.50 & -1.22\\  

        \gls{ours} (all epsilon) & \textbf{2.50} & \textbf{1.12} & \textbf{-1.38}\\  

        \bottomrule
    \end{tabular}
\end{table*}

\end{document}